\documentclass[sn-apa]{sn-jnl}

\usepackage{graphicx}%
\usepackage{multirow}%
\usepackage{amsmath,amssymb,amsfonts}%
\usepackage{amsthm}%
\usepackage{mathrsfs}%
\usepackage[title]{appendix}%
\usepackage{xcolor}%
\usepackage{textcomp}%
\usepackage{manyfoot}%
\usepackage{booktabs}%
\usepackage{algorithm}%
\usepackage{algorithmicx}%
\usepackage{algpseudocode}%
\usepackage{listings}%
\usepackage{caption}
\usepackage{subcaption}
\usepackage{color,soul}

\theoremstyle{thmstyleone}%
\theoremstyle{thmstyletwo}%

\theoremstyle{thmstylethree}%

\begin{document}

\title[Article Title]{On the Dynamics of Mating Preferences in Genetic Programming}


\author*{\fnm{José Maria} \sur{Simões}* }\email{josecs@dei.uc.pt}

\author{\fnm{Nuno} \sur{Lourenço}}\email{naml@dei.uc.pt}

\author{\fnm{Penousal} \sur{Machado}}\email{machado@dei.uc.pt}

\affil{\orgdiv{CISUC}, \orgname{Department of Informatics Engineering, University of Coimbra}, \orgaddress{\street{Polo II - Pinhal de Marrocos}, \city{City}, \postcode{3030}, \state{Coimbra}, \country{Portugal}}}


\abstract{Several mating restriction techniques have been implemented in Evolutionary Algorithms to promote diversity. From similarity-based selection to niche preservation, the general goal is to avoid premature convergence by not having fitness pressure as the single evolutionary force. In a way, such methods can resemble the mechanisms involved in Sexual Selection, although generally assuming a simplified approach. Recently, a selection method called mating Preferences as Ideal Mating Partners (PIMP) has been applied to GP, providing promising results both in performance and diversity maintenance. The method mimics Mate Choice through the unbounded evolution of personal preferences rather than having a single set of rules to shape parent selection. As such, PIMP allows ideal mate representations to evolve freely, thus potentially taking advantage of Sexual Selection as a dynamic secondary force to fitness pressure.
However, it is still unclear how mating preferences affect the overall population and how dependent they are on set-up choices.
In this work, we tracked the evolution of individual preferences through different mutation types, searching for patterns and evidence of self-reinforcement. Results suggest that mating preferences do not stand on their own, relying on subtree mutation to avoid convergence to single-node trees. Nevertheless, they consistently promote smaller and more balanced solutions depth-wise than a standard tournament selection, reducing the impact of bloat. Furthermore, when coupled with subtree mutation it also results in more solution diversity with statistically significant results.}




\maketitle

\section{Introduction}\label{sec:introduction}

Diversity maintenance is paramount in Genetic Programming (GP), as it is for most Evolutionary Algorithms (EA) \citep{Eiben2015,Poli2008,Burke2004,Hien2006}. It is well known that such algorithms are prone to losing general diversity within a given set of solutions as the evolutionary process unfolds, a phenomenon often termed premature convergence \citep{Eiben2015}. This particular phenomenon hinders the search process, having the undesired potential of condensing all solutions around a single and sub-optimal one. Formally, this is often the result of an imbalance between exploitation (improving results in a given region of the search space) and exploration (promoting a broad search for new regions) to the detriment of the latter \citep{ConvergenceEiben98, AbdelBasset2018}. 

There is a vast literature on approaches to prevent premature convergence, most commonly falling under the umbrella of explicit approaches such as Fitness Sharing \citep{Goldberg1987}, Crowding \citep{Jong1975, Mahfou1992}, and some implicit ones such as Island Models or Cellular EAs \citep{Eiben2015}. Essentially, whether explicitly or implicitly, these methods establish some type of selection or mating restrictions to preserve niches that can be maintained for isolated exploitation. Just as the genesis of EAs, most of these methods are derived from natural phenomena, whether by treating fitness as a finite resource or by isolating sub-populations \citep{Eiben2015, Glibovets2013, Miller1996}. Selection based on individual fitness instances has also been explored, particularly through Lexicase Selection \citep{Lexicase_Original, Lexicase_Original_Two}, where parents are chosen based on their performance on specific instances (or objectives) rather than an overall aggregated fitness measure. This approach has been utilized across various types of evolutionary computation, including symbolic regression in GP  \citep{Lexi_Regression} and multi-objective optimization \citep{Lexi_Multi_Obj}, having shown good results in both performance and diversity maintenance \citep{Lexi_diversity}.

Other approaches follow some alternative restriction mechanisms that in turn can be more unpredictable, at least to some extent. Expanding on the foundation of Natural Selection and survival of the fittest -- as described by Charles Darwin \citep{Darwin1} --, some authors have focused mainly on the selection stage at the individual level rather than sticking to a general fitness or population view. Some resemble similarities to Sexual Selection and Mate Choice although some are not explicitly described as such. Sexual Selection, also pioneered by Darwin \citep{Darwin1, Darwin2}, can be described as a parallel force to Natural Selection, which has the potential to shape evolutionary paths and establish differences between organisms of the same species \citep{gayon2010, Alonzo2019, Brock2007, Ralls2009}. Mainly divided into two categories, we can observe the influence of Sexual Selection via competition for mates or by conveying evolutionary advantage to some traits critical to be accepted as a mate (i.e., Mate Choice) \citep{Jones2009}. Since its early formulation, Sexual Selection gradually became a growing field of interest and today we know it holds an important role in the visible diversity between and within species \citep{Hollocher2013}. However, with scientific progress came the notion that the dynamics underlying Sexual Selection are more diverse and complex than could be perceived in Darwin's days \citep{Ralls2009, Fisher1930, Zahavi1975}. 

Usually, the transposition of the concept of Sexual Selection to EAs is mostly based on broader notions concerning sexual reproduction. In Genetic Algorithms we find studies implicitly applying two different sexes \citep{Drezner2006, Zhu2006, Srinivas1994}, multiple genders \citep{Vrajitoru2008} or even gender separation coupled with different selective pressures \citep{Omori2005, Cheng2012} and dissimilar gendered mate choice \citep{Varnamkhasti2012, Jalali2012}. Although scarcer, some Sexual Selection mechanisms applied to GP can also be found, where authors have also studied the impacts of having dissimilar mates \citep{Fry2005} or self-adapting mate selection functions \citep{Tauritz2007}. In a recent work by Leitão et al. \citep{Leitao2019, Leitao2013, Leitao2015}, Sexual Selection was deliberately applied to GP by mimicking preferences for an ideal mate. The authors proposed the Mating Preferences as Ideal Mating Partners in the Phenotype Space model (PIMP) targeting symbolic regression as test cases, in which partners are chosen based on the representation of an ideal mate. As argued by the authors, this model has the potential of establishing a secondary selective force that may act against Natural Selection (i.e., fitness pressure), therefore promoting more diversity within the evolving population. To our knowledge, this is among the most detailed and in-depth works on Mate Choice mechanisms applied to GP. 

Compared to a standard tournament selection, the framework showed performance improvements in several symbolic regression problems, also pointing towards a more exploratory search. Moreover, the authors provided an extensive analysis of the dynamics that emerge throughout the evolutionary process with some interesting findings: automatic segregation within the population (resembling sexual differentiation), differences between individuals in each role (Chooser vs. Courter), or even the fact that mating preferences can evolve against or in favour of fitness pressure.  

The work provided by \cite{Leitao2019} points towards an interesting potential behind the dynamics of Mate Choice applied to EAs, especially as a complementary force to aid diversity. While restrictions such as assortative or disassortative mating are, in a way, more predictable (in the sense that the structure used to compare mates is strictly connected to fitness), ideal mate representations add another dimension to the evolutionary process, being free to evolve alongside or against Natural Selection. 
To our knowledge, there is currently no in-depth research on how sexual selection and mate choice evolve, particularly in the context of GP, since many studies tend to focus primarily on performance. While we acknowledge performance is important, we believe understanding mating preferences is essential for uncovering their advantages and limitations, specially when planning the algorithm to use beforehand. As such, in this work, we study how mating preferences evolve, how their dynamics relate to different population aspects, and ultimately their ability to act as a self-reinforced phenomenon.  
For that, we use different metrics to those applied in the original work of PIMP \citep{Leitao2019, Leitao2013}, as well as different mutation methods, tracking preferences in isolation in order to have a general view of how they operate. Three different symbolic regression test cases are used to study thoroughly the side effects of mate choice.

Results suggest that mate choice implemented as it is in PIMP fails to reach a sustainable evolutionary path independently. More precisely, we found that, without a mutation that promotes tree growth, preferences tend to converge to the simplest structure possible. Further analysis shows that such convergence is unlikely to be caused by a self-reinforcement process alone, meaning that the PIMP method is quite sensitive to the mutation method used. However, results also show that it still manages to create role segregation, promoting smaller and more diverse trees than a standard tournament selection with statistically significant differences. Finally, a more complex problem was also used where a Diabetes dataset \citep{2020SciPy} proved to be hard to tackle by both approaches and where PIMP did not outperform a standard GP with tournament selection. Nonetheless, it still promoted a more balanced tree depth distribution and more unique solutions with statistically significant differences, which in turn might reduce the undesired side-effects of bloat (when solutions grow unnecessarily large), a major issue often found in GP when dealing with real-world problems \citep{Poli2008, Luke2006, Trujillo2016, Angelis2023}. 

As such, in this work, we clarify the dynamics of mating preferences in GP. The main contributions in this work are: 
\begin{itemize}
  \item The theory behind preference reinforcement does not hold under specific mutation conditions.
  \item Mating preferences as a secondary evolutionary force encourage a greater variety of solutions.
  \item Mating preferences also seem to decrease the complexity of evolutionary trees.
\end{itemize}

The remainder of this article is structured as follows: we first provide an overview of the PIMP method along with its previous findings and our motivation behind this study. We then present the methodology used in our experiments, followed by the results and a discussion on the gathered data and future considerations.

\section{PIMP}\label{sec:PIMP}

In this section, we briefly describe the architecture and general mechanisms behind the PIMP framework. Furthermore, we provide a summary of the results presented by the authors, which set the stage for the remainder of our work.
As we aim at understanding better how mating preferences evolve, PIMP has a central role in our experiments. Therefore, we provide an overview of the method as well as a concrete explanation of why we believe it to be relevant to have a closer look at the dynamics of mating preferences in the first place. 

\subsection{Architecture}
\begin{figure}[t]
  \centering
  \includegraphics[width=2.5in]{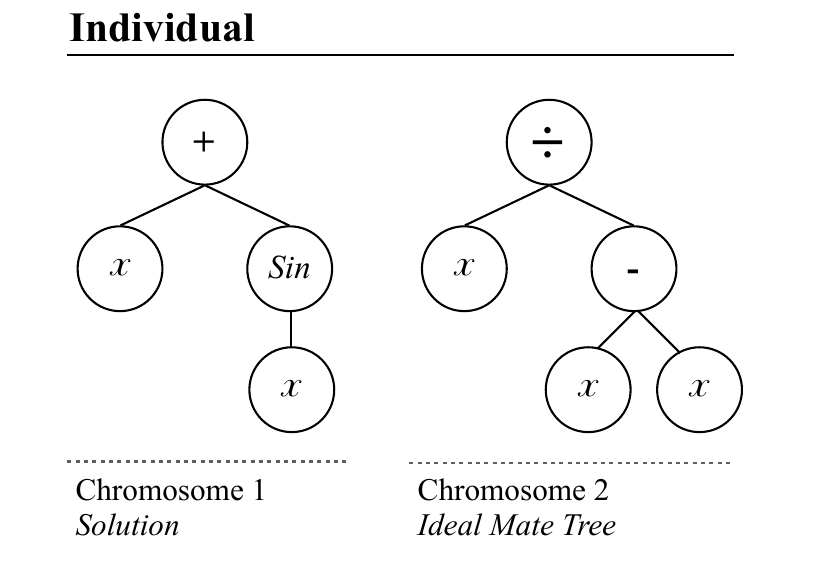 }
  \caption{Illustration of the composition of an individual under the PIMP approach.}
  \label{imagem_representation}
\end{figure}

PIMP was proposed by Leitão et al. \citep{Leitao2019, Leitao2013, Leitao2015} as a way of modelling mating preferences. The method aims at transposing the phenomenon of  Mate Choice, which often occurs when potential mates are chosen based on secondary sexual characteristics \citep{Richards2011, andersson2006}. As Sexual Selection itself has the potential to promote diversity, the authors propose that the same notions can be applied to EAs. The framework is applied to GP, thus our experiments apply the same principles under a GP representation. Arguably one of the most appealing factors of PIMP is that it is quite straightforward to implement. The method differs from a Standard GP approach (i.e., both parents are drawn from a simple fitness-based selection) at two main levels: individuals and selection. Individuals are composed of two chromosomes: the first chromosome is the solution to any given problem (which in turn establishes the fitness of the individual in its standard form); the second chromosome holds the preference, and it encodes a representation of the ideal mate. Both chromosomes can be built from the same function set, which practically makes it easier to implement. Fig. \ref{imagem_representation} shows an example of the structure of an individual in PIMP. 

Regarding parent selection (see Algorithm \ref{PIMP_alg}), the first individual (first parent) is selected via a standard tournament with the default size of 5 (becoming the Chooser) where the first chromosome is used to measure its fitness. Then, a set of individuals is selected at random -- the potential mates, or Courters. At this stage, the second chromosome of the Chooser is activated and compared against the first chromosome of each potential mate. The Courter having the closest solution to the preference of the Chooser is selected as a mate. This comparison can be measured in the same way as solutions are compared against the objective function -- for instance, if the mean squared error is being used to compute fitness, this same metric can be used to compare candidates against preferences. In practice, this means that preferences are not directly compared to candidate trees, but rather their fitness. Although this indirect comparison does not convey too much of additional computational costs, it means that different trees can be translated into a perfect match of preference and candidate. 

\begin{figure}[h]
\centering
 \includegraphics[width=4.7in]{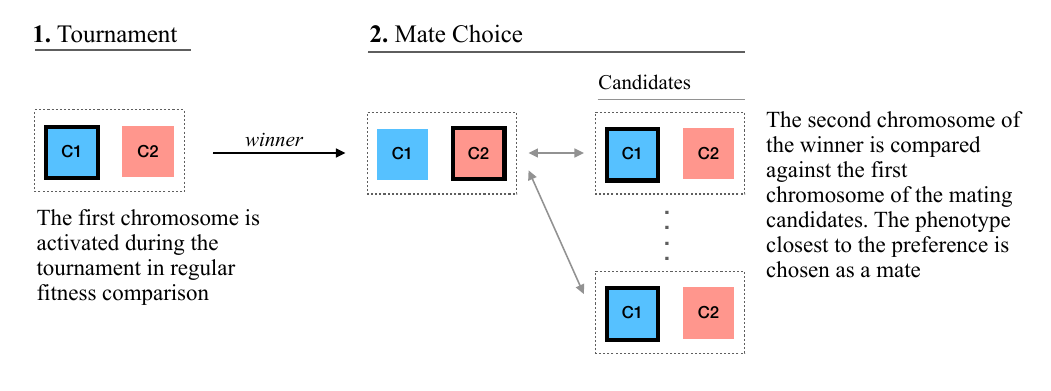}
 \caption{Illustration of the selection scheme in PIMP.}
 \label{selection}
\end{figure}

Fig. \ref{selection} illustrates the selection phase. In a way, one might say that the Chooser has a version of the ideal solution that it would prefer to mate with. Note that PIMP determines no roles, meaning that an individual can act as a Courter or as a Chooser in the same generation. 

Finally, both chromosomes have the same chance of being recombined (where preferences can only be recombined with preferences and solutions with solutions) and mutated.

\begin{algorithm}
	\caption{PIMP parent selection method}
        \label{PIMP_alg}
	\begin{algorithmic}[1]
		\While {$selected\_set\_size< pop\_size$}
			 \State {$parent1 = tournament(pop)$}
              \State {$candidates\_set = random(pop)$}
              \State {$evaluate\_candidates(parent1, candidates\_set)$}
              \State {$parent2 = select\_best(candidates\_set)$}
              \State {$selected\_set.append(parent1, parent2)$}
            \EndWhile
        \end{algorithmic} 
\end{algorithm}

\subsection{Kown Effects and Motivation}
In the original work on PIMP, the method was tested against a Standard GP Approach (i.e., both parents chosen through a regular tournament) on 52 symbolic regression instances and showed performance gains, reaching statistical significance on more than half when mutation was introduced. Furthermore, the authors reported evidence that pointed towards diversity gains when using PIMP, arguably one of the main advantages of using the technique. Nonetheless, we believe that some points should be briefly addressed that set the stage for our study. 


In PIMP, although Choosers compete in a tournament, the results often showed more diversity among individuals under fitness pressure when compared to a standard tournament selection. As argued by the authors, this might be a result of having only one parent selected through fitness. Moreover, Courters are merely regulated by the existing preferences among the current Choosers' pool. 
This means that the force guiding the Courters is rather unpredictable (especially in early generations), thus potentially resulting in an unbounded evolutionary path. 
This also means that as generations go by, Choosers are subject to different pressures (even in the same generation), as long as preferences do not converge. In fact, there is evidence that preferences tend to lose diversity \citep{Leitao2019}, which can be seen as a self-reinforcing process of the Chooser’s preferences, resembling a Fisherian process (that is, the evolution of specific traits due to Mate Choice) \citep{Fisher1930}. Still, even if preference convergence happens, this has the potential of establishing a force contrary to that of Natural Selection, thus being able to maintain a more diverse set of individuals. Nonetheless, it was pointed out that on average preferences tend to become similar as the evolutionary process goes on. Therefore, we believe it to be important to understand why this happens. 

Another interesting point in the original study is that of role (or sex) segregation. By applying different and potentially divergent forces when selecting parents, PIMP promotes automatic role segregation, dividing the population mostly into Choosers and Courters, while a smaller fraction behaves as both. This ‘specialisation’ also culminates in structural differences in roles. As pointed out by the authors, Courters showed higher levels of entropy than Choosers (resulting from the random initialisation of preferences), and eventually preferences seemed to promote some kind of negative assortative mating, where Courters tend to evolve to simpler structures while Choosers display more complex ones. As such, we set out to study how this impacts the evolutionary process at a population level. 

Although in nature mating preferences can be hard to examine or to track, in algorithmic practice we can look closer into them. While performance is often the most important aspect of an EA, understanding whether the nature-inspired phenomena hold true is also relevant from a theoretical standpoint, which in turn can have practical implications. With this, we believe that there’s room to perform a deeper study on mating ideals, not only due to the promising results found in the past but also considering the inherently complex dynamics of Mate Choice and, specifically, mating preferences. Given that preferences are the crucial factor that characterises PIMP, we believe that focusing primarily on them can help answer what might be happening at this level. A recent study shows that this same method behaves differently from a Random Mate approach \citep{SimoesEvostar}, which we believe stresses the importance of the proposed question in this study: How do mating preferences evolve in GP? 

As we will discuss further in this document, this central question unfolds into other important aspects that should also be examined, such as the importance of role segregation and the impact these different roles have on overall performance. Finally, we must state that other diversity preservation approaches were left out of these experiments as our goal is not to compare this method against existing ones, but rather to assess whether mating preferences are sustainable and the effects they have on the population.

\section{Methodology}
We deliberately took an explorative approach to study how preferences evolve under this framework. Having original work on PIMP as a starting point \citep{Leitao2019, Leitao2013}, we decided to examine several aspects that were not covered there, such as tree structure, diversity of preferences and the impact of different mutation variations. From that, we nonetheless experimented with different mutation methods while analysing the evolution of preferences in general. To analyse the gathered data, we established different metrics: 
\begin{itemize}
\item Mean Best Fitness - From the best fitness at the last generation from each run.
\item Average Unique Trees - Percentage of unique trees in a population in a given time.
\item Average Depth - Average depth of the trees in each run.
\end{itemize}

Three symbolic regression functions were used as testing instances: Koza-1, Nguyen-6 and Pagie-1, specified in Table \ref{funcions}. All these share the same function set (detailed in Table \ref{table_function_set}) (a summary of these instances can be found in \citep{McDermott2012}). Furthermore, the Scipy Diabetes \citep{2020SciPy} dataset was also included when testing against a standard tournament selection. 
For each algorithm or variation, the results presented are gathered from sets of 30 runs using shared seeds.

\begin{table}[b]
\caption{Objective functions used.}\label{funcions} 
\renewcommand{\arraystretch}{1.5}
\begin{tabular}{ccc}
\toprule
Name & Vars & Objective Function \\ \midrule
Koza-1 & 1 & $x^{4}+x^{3}+x^{2}+x$ \\
Nguyen-6 & 1 & $\sin(x)+\sin(x+x^2)$ \\
Pagie-1 & 2 & \large{ $\frac{1}{1+x^{-4}} + \frac{1}{1+y^{-4}}$ } \\ 
\bottomrule
\end{tabular}
\end{table}

\begin{table}[b]
\caption{Function Set (same for all instances).}\label{table_function_set}
\renewcommand{\arraystretch}{1.5}
\begin{tabular}{cc}
\toprule
Functions & Constants \\ \midrule
\multicolumn{1}{c}{$+$, $-$, $\times$, $\%$, $\sin$, $\cos$, $e^n$, $\ln(|n|)$}  & \multicolumn{1}{c}{None} \\ 
\bottomrule
\end{tabular}
\end{table}

We chose three symbolic regression instances that were part of the original work \citep{Leitao2019}, hence our experiments can be seen as complementary to extend the existing results. As such, the set-up for our experiments was kept mostly the same as that employed originally. Nonetheless, we must highlight some minor changes. Firstly, all runs evolved during 1500 generations (rather than 500) to observe the behaviour in the long run. Second, we decided to allow crossover and mutation to act on the root node. As our experiments went on, different mutation versions were used in an experimental set-up. Mainly, Subtree mutation was used (using the method Ramped Half-and-half to generate trees with a random size between 2 and 7) and Node Replacement (where each node has a 5\% chance of being changed whenever an individual goes through mutation). One-point crossover was used to exchange subtrees between parents (preference chromosomes were also subject to crossover). Finally, we have used a standard tournament selection (each parent is selected through a tournament of size = 5) to assess performance gains. The general set-up used for all algorithms is presented in Table \ref{Set_up}. Mean squared error was used to compute fitness and to measure mating candidates (i.e., the preference function of the Chooser is compared to the solution chromosome of the Courters).

Although a great part of this article is focused on diversity and tree structure, mean best fitness (MBF) is also included, being used as a standard way of measuring algorithmic performance. Regarding diversity, Tree Edit Distance is too expensive computation-wise, therefore it was decided to count the unique trees as a baseline: if a specific algorithm or set-up fails to produce enough unique trees, then convergence is more likely to occur. Tree depth is included as the main way of comparing tree structures and complexity. Finally, in the remainder of this document, we identify Choosers as individuals that were only selected via tournament, while Courters as individuals that were exclusively selected by a Chooser. 

\begin{table}[h]
\caption{General Set up.}\label{Set_up}
\begin{tabular}{@{}lll@{}}
\toprule
 & Population Size & 100 \\
General Parameters & Generations & 1500 \\
 & Elitism & None \\ \midrule
Individual Builder & Ramped half-and-half & random(2,7) \\ \midrule
 & Crossover Prob. & 0.9 \\
Breeding & Mutation Prob. & 0.05 \\
 & Max Depth & 17 \\ \bottomrule
\end{tabular}
\end{table}

\section{Results}
In this section we present the results gathered from our different experiments. It is divided into three subsections, where we begin by exploring the evolution and general structure of preferences, followed by its main effects on the population. Finally, we establish a comparison against a standard tournament selection to assess potential gains of the method regarding diversity, tree depth and performance.

\subsection{How do preferences evolve?}
The first experiment in our study aims at understanding the evolution of preferences using PIMP with a subtree mutation. For comparison purposes, we tracked the evolution of the percentage of unique solutions and preferences separately. The average results from 30 runs at termination are presented in Table \ref{PIMP_subtree_unique} (the standard deviation for each value is given in parentheses). Based on this data, we notice that solutions hold high levels of diversity and preferences hold a slightly lower percentage overall. Nonetheless, both seem to sustain a considerable amount of unique trees. 

\begin{table}[b]
\caption{Average Unique Trees at termination (PIMP w/ Subtree Mutation).}\label{PIMP_subtree_unique}
\renewcommand{\arraystretch}{1.5}
\begin{tabular}{cccc}
\toprule
\textit{} & Koza-1 & Nguyen-6 & Pagie-1 \\ \midrule
Solution Chromosome & 86\%{\footnotesize ($\pm$6)} & 84\%{\footnotesize ($\pm$5)} & 90\%{\footnotesize ($\pm$5)} \\
Preference Chromosome & 79\%{\footnotesize ($\pm$7)} & 78\%{\footnotesize ($\pm$9)} & 87\%{\footnotesize ($\pm$7)} \\ \bottomrule
\end{tabular}
\end{table}

\begin{figure}[ht]
  \centering
  \includegraphics[width=3.8in]{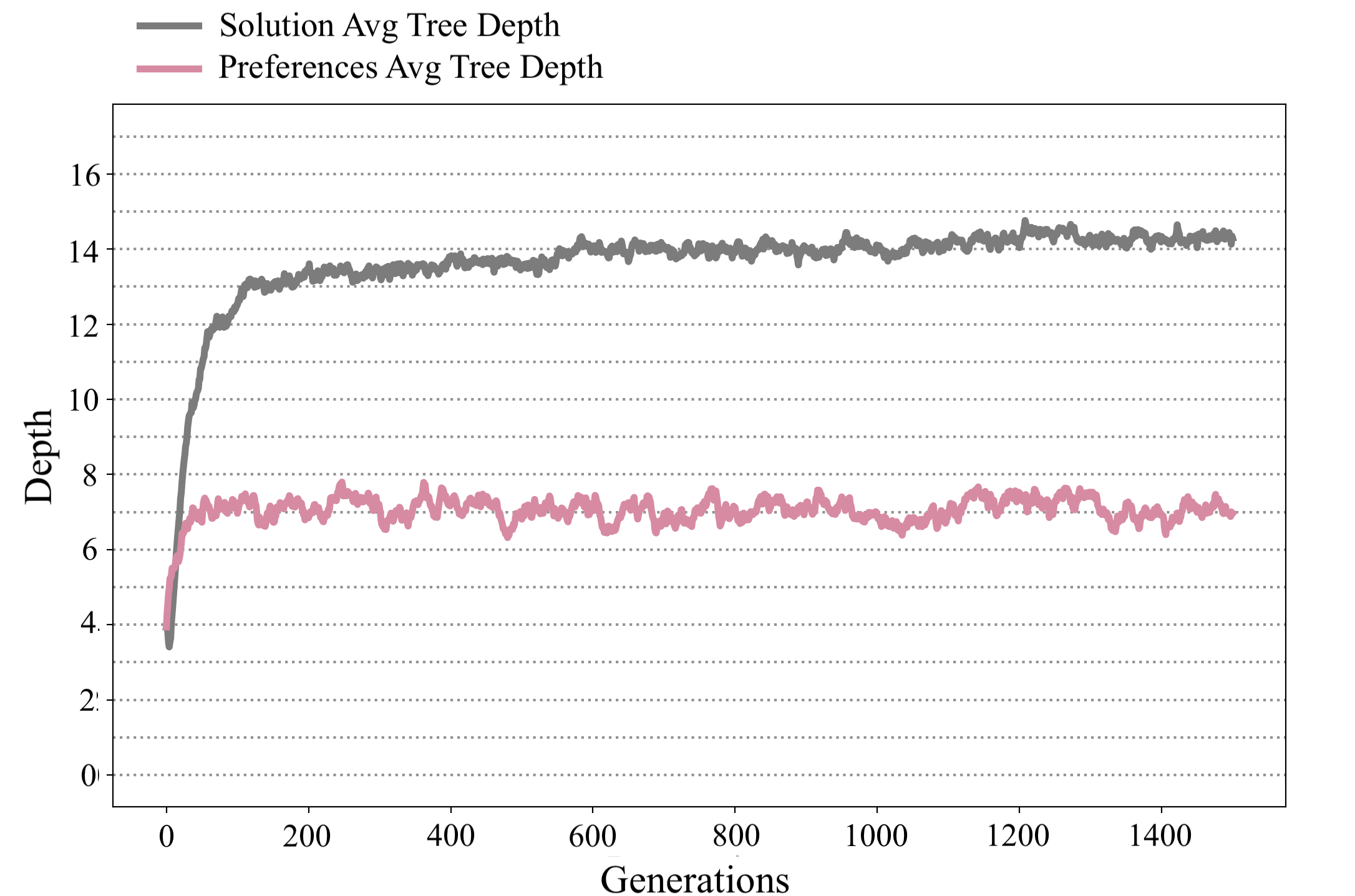}
  \caption{PIMP w/ Subtree mutation Average Tree Depth - Koza-1.}
  \label{subtree_depth_koza1}
\end{figure}

Considering this, there is a relevant point that should be addressed: are there structural differences between solutions and preferences? For that, we tracked the overall depths of each group, which tells us the potential complexity involved in each.
These results are presented in Table \ref{PIMP_subtree_avg_tree_depth}, with Fig. \ref{subtree_depth_koza1} serving as an illustration of the evolutionary dynamics of the average depths. While a depth of 17 implies 17 levels of depth (the defined maximum), a depth of zero means that a given tree has only one node. Interestingly, we observe a  difference in the main average depths of the groups: preferences tend to be shallower than solutions, globally converging towards roughly half the depth of the latter.

\begin{table}[ht]
\caption{Average Tree Depth at termination (PIMP w/ Subtree Mutation). }\label{PIMP_subtree_avg_tree_depth}
\renewcommand{\arraystretch}{1.5}
\begin{tabular}{cccc}
\toprule
\textit{} & Koza-1 & Nguyen-6 & Pagie-1 \\ \midrule
Solution Chromosome & 14.3{\footnotesize ($\pm$1.4)} & 14.2{\footnotesize ($\pm$1.0)} & 14.5{\footnotesize ($\pm$1.2)} \\
Preference Chromosome & 6.9{\footnotesize ($\pm$1.7)} & 7.2{\footnotesize ($\pm$1.9)} & 7.1{\footnotesize ($\pm$1.8)} \\ 
\bottomrule
\end{tabular}
\end{table}

This behaviour is not completely new or unexpected (in the original work, the authors pointed out likely differences in complexity), yet following the same line of thought this can be driven by two different causes (not necessarily mutually exclusive): 1) Preferences converge to smaller and thus less complex trees merely due to selective force, posing some kind of evolutionary advantage (e.g., having preferences for smaller mates translates into fitness gains); or 2) Preferences assume less complex structures due to the absence of a strong selective force, potentially preventing preferences from evolving freely in a self-reinforced fashion -- this might be explained by smaller trees being less affected by destructive crossover and also by the Crossover Bias Theory (CBT), which states that subtree crossover generates a large number of small trees in the early generations \citep{Poli2007, Trujillo2019}.  
If the latter holds true, then preferences will always have a tendency to decrease in average depth and complexity as the evolution unfolds. Curiously, in our results, we observe that preferences tend to converge to the maximum limit as defined by the current mutation set up as a restriction (see Table \ref{Set_up}). 

This second hypothesis can be tested by running the same instances without any mutation method. As such, another set of experiments was performed under the same conditions yet allowing only crossover. The results regarding tree depth are summarized in Table \ref{PIMP_noMutation_avg_tree_depth}.

\begin{table}[b]
\caption{Average Tree Depth at termination (PIMP without Mutation). }\label{PIMP_noMutation_avg_tree_depth}
\renewcommand{\arraystretch}{1.5}
\begin{tabular}{cccc}
\toprule
\textit{} & Koza-1 & Nguyen-6 & Pagie-1 \\ \midrule
Solution Chromosome & 10.8{\footnotesize ($\pm$5.7)} & 10.6{\footnotesize ($\pm$5.8)} & 13.5{\footnotesize ($\pm$2.7)} \\
Preference Chromosome & 0{\footnotesize ($\pm$0.0)} & 0.4{\footnotesize ($\pm$1.4)} & 0{\footnotesize ($\pm$0.0)} \\ \bottomrule
\end{tabular}
\end{table}

\begin{figure}[h]
 \centering
 \begin{subfigure}{0.49\textwidth}
     \includegraphics[width=\textwidth]{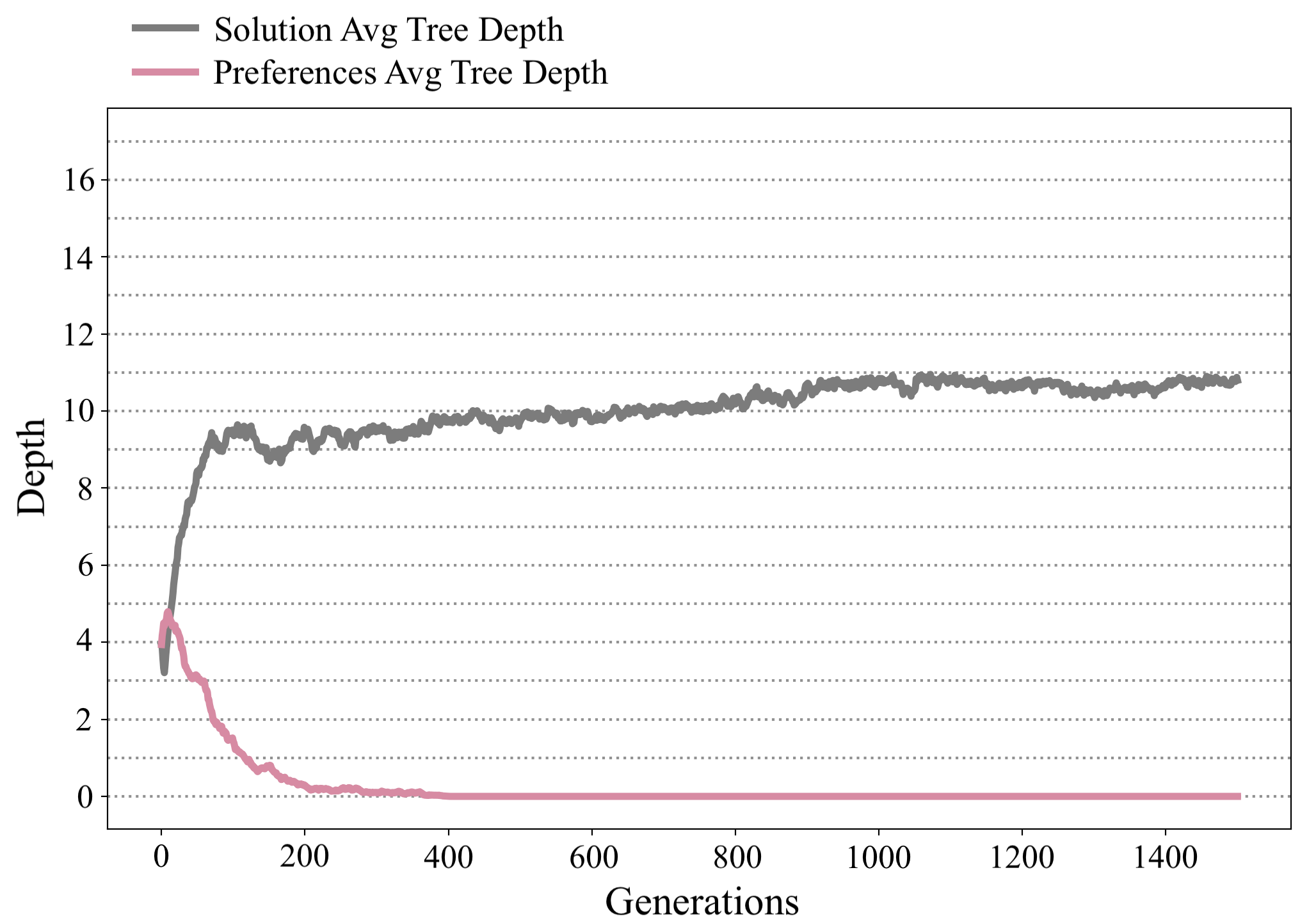}
     \caption{Koza-1.}
     \label{subtree_depth_nomut_koza1}
 \end{subfigure}
 \hfill
 \begin{subfigure}{0.49\textwidth}
     \includegraphics[width=\textwidth]{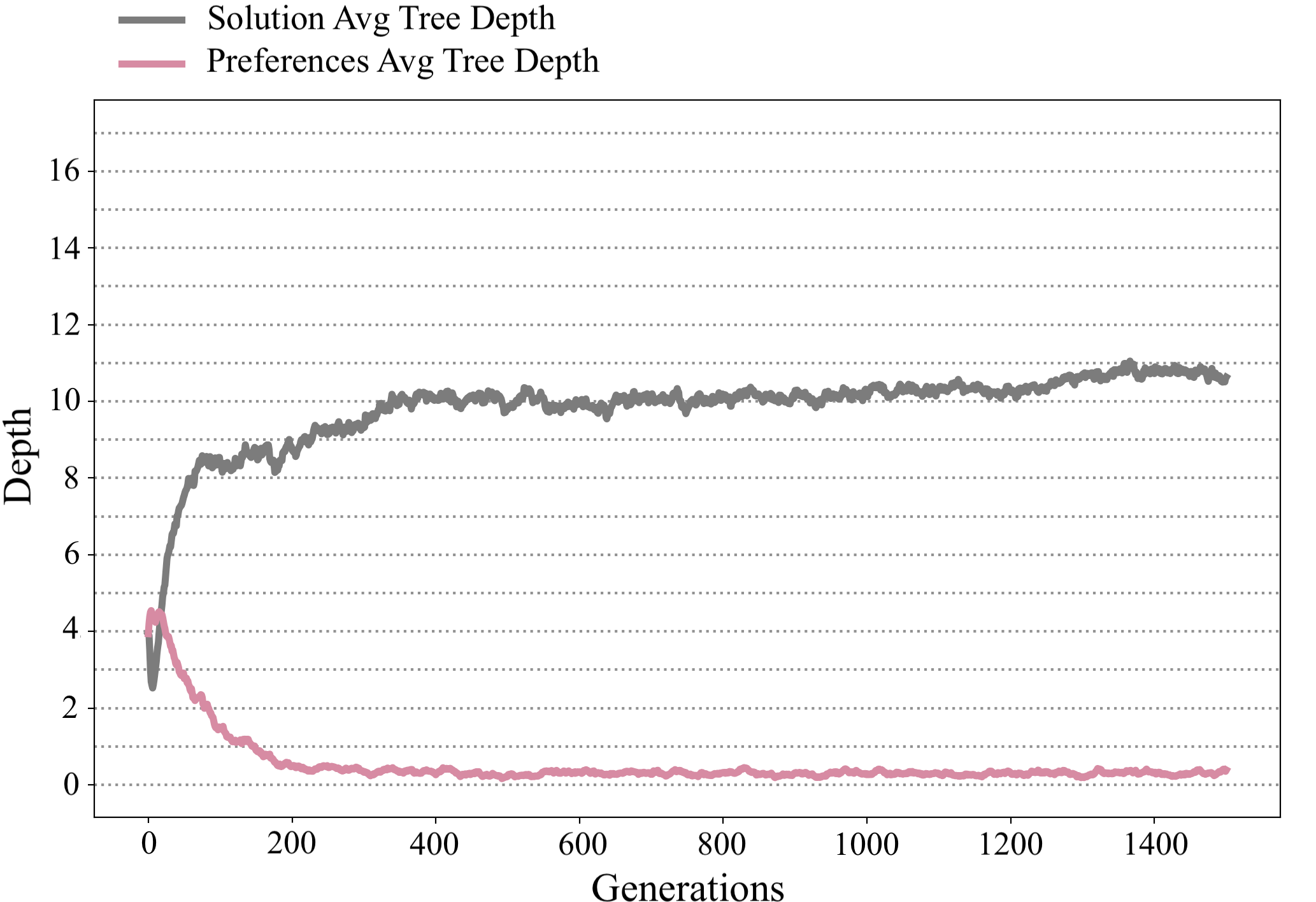}
     \caption{Nguyen-6.}
     \label{subtree_depth_nomut_nguyen6}
 \end{subfigure}
 \hfill
 \begin{subfigure}[b]{0.49\textwidth}
     \centering
     \includegraphics[width=\textwidth]{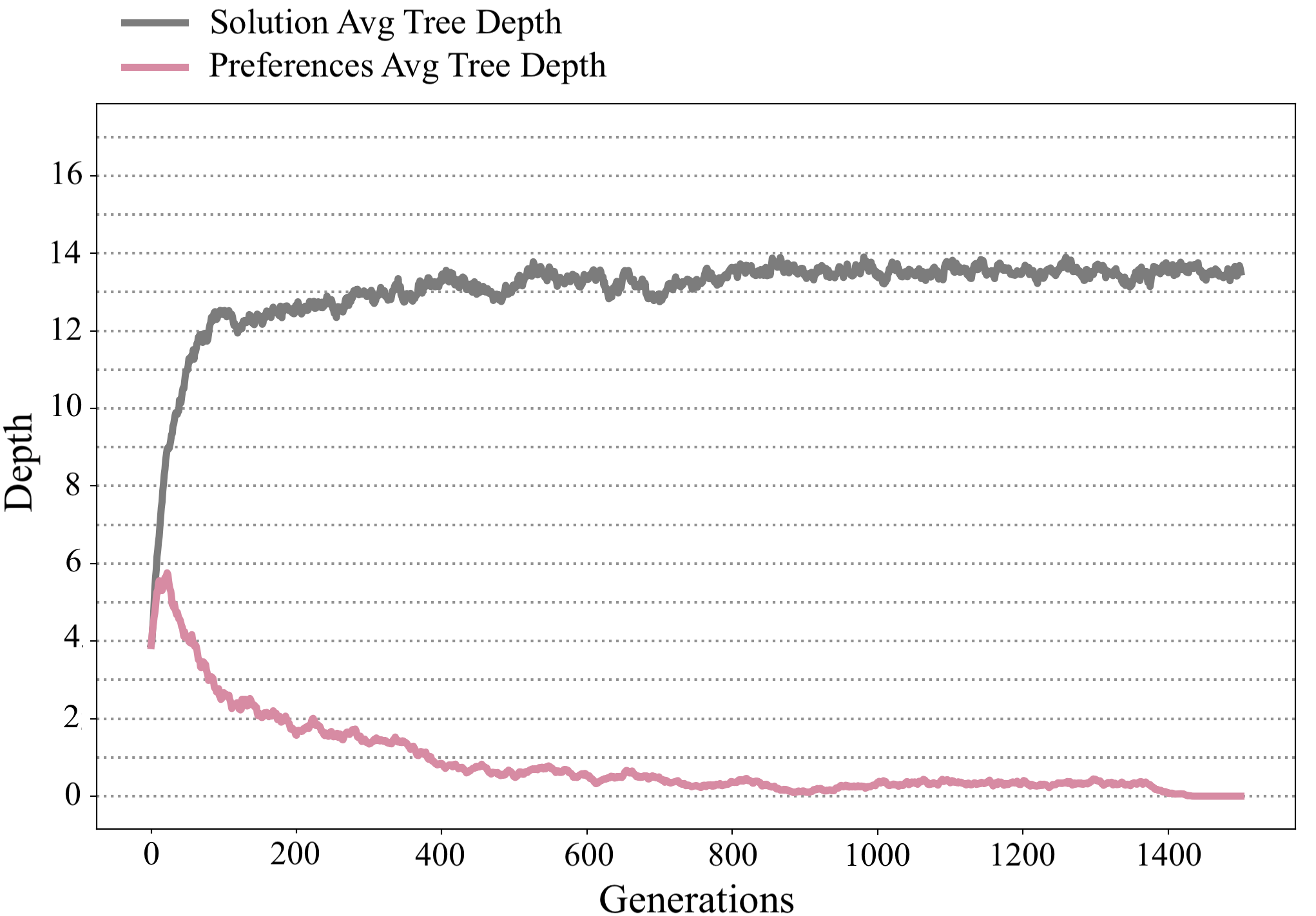}
     \caption{Pagie-1.}
     \label{subtree_depth_nomut_pagie1}
 \end{subfigure}
    \caption{PIMP without mutation Average Tree Depth.}
    \label{fig:three graphs}
\end{figure}

Here we observe that preferences converge in fact to the smallest depth possible, meaning that at termination the average preference chromosome had only one node (i.e., the root node). Furthermore,  Fig. \ref{subtree_depth_nomut_koza1}, \ref{subtree_depth_nomut_nguyen6} and \ref{subtree_depth_nomut_pagie1} show that this convergence happens quite early and drastically in the evolutionary process. Simultaneously, we also observe that solutions tend to assume a slightly smaller overall depth, which in turn might be explained by the general preference for smaller mates.


\begin{table}[h]
	\renewcommand{\arraystretch}{1.5}
	\caption{Average Tree Depth at termination (PIMP w/ Node Replacement Mutation). }\label{PIMP_nodeReplace_avg_tree_depth}
	\begin{tabular}{cccc}
		\toprule
		\textit{} & Koza-1 & Nguyen-6 & Pagie-1 \\ \midrule
		Solution Chromosome & 10.1{\footnotesize ($\pm$5.6)} & 10.1{\footnotesize ($\pm$6.0)} & 13.3{\footnotesize ($\pm$3.3)} \\
		Preference Chromosome & 0{\footnotesize ($\pm$0.0)} & 0{\footnotesize ($\pm$0.0)} & 0{\footnotesize ($\pm$0.0)} \\ \bottomrule
	\end{tabular}
\end{table}

Having evidence that suggests that CBT might be stronger than a natural reinforcement of the preferences, we conducted the same experiment with another form of mutation: Node Replacement. This allows preferences to be more diverse and potentially find a stability point for self-reinforcement without affecting its overall depth. After running PIMP with this different mutation for 30 runs, we observe that the results displayed in Table \ref{PIMP_nodeReplace_avg_tree_depth} are quite similar to that of no mutation, with the same convergence happening early on in the evolution.

\begin{figure}[t]
  \centering
  \includegraphics[width=3.5in]{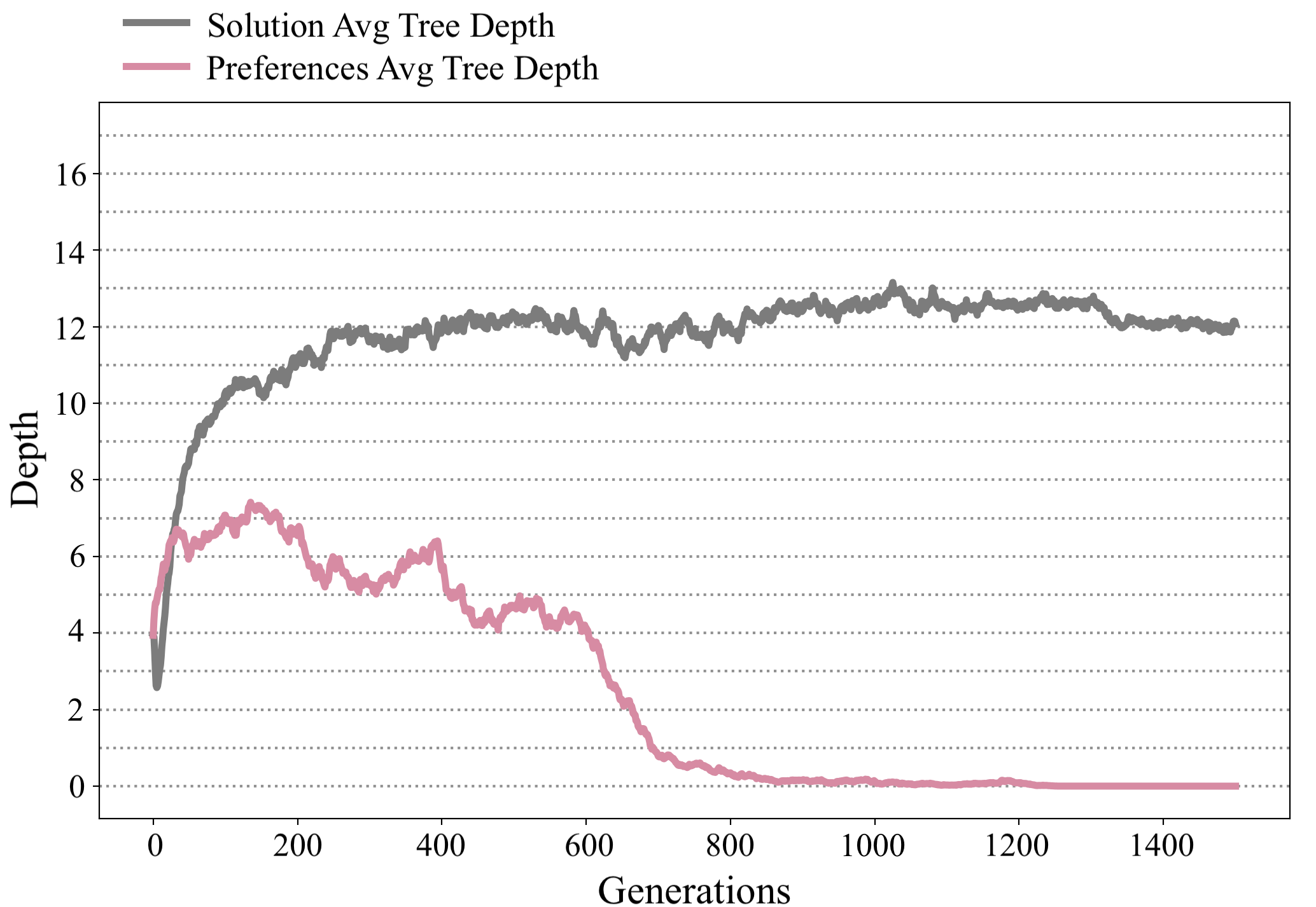}
  \caption{PIMP w/ Hybrid mutation Average Tree Depth - Nguyen-6.}
  \label{subtree_depth_hybrid_Nguyen6}
\end{figure}

Although these results seem to be in line with the idea that preferences fail to sustain in the absence of a mutation that potentially increases tree depth (like the subtree mutation used in our experiments), there is also one last point regarding tree depth that should be considered: are preferences under Subtree mutation strong enough to develop a self-reinforcement factor that sustains its overall depth? To answer this question, we conducted another experiment where the mutation growth limit changed as generations went by, eventually allowing only mutation on a single node, mostly in a hybrid-mutation scheme fashion. The activation sequence is detailed in Table \ref{hybrid_description}. This scheme allows us to understand better whether preferences have developed a sustainable force by the time Node Replacement mutation is introduced. As we can observe from Table \ref{PIMP_hybrid_avg_tree_depth}, the results also show that preferences converge to a single node, and Fig. \ref{subtree_depth_hybrid_Nguyen6} illustrates precisely this: preferences seem to struggle to maintain their higher depth levels in the absence of a growth method. Furthermore, it is visible that a drastic convergence to single-node preferences happens around generation 600 - precisely when Node Replacement mutation is introduced.

\begin{table}[b]
\caption{Hybrid Mutation Description.}\label{hybrid_description}
\renewcommand{\arraystretch}{1.5}
\begin{tabular}{ccccc}
\toprule
Gen & 0-200 & 200-400 & 400-600 & 600-1500 \\ \hline
Mutation Method & Subtree & Subtree & Subtree & Node Replacement \\ \midrule
Grow Min/Max & 2/6 & 2/4 & 2/2 & None \\ \bottomrule
\end{tabular}
\end{table}

\begin{table}[b]
\renewcommand{\arraystretch}{1.5}
\caption{Average Tree Depth at termination (PIMP Hybrid Mutation). }\label{PIMP_hybrid_avg_tree_depth}
\begin{tabular}{cccc}
\toprule
\textit{} & Koza-1 & Nguyen-6 & Pagie-1 \\ \midrule
Solution Chromosome & 14{\footnotesize ($\pm$1.9)} & 12.1{\footnotesize ($\pm$5.3)} & 14.3{\footnotesize ($\pm$1.3)} \\
Preference Chromosome & 0{\footnotesize ($\pm$0.0)} & 0{\footnotesize ($\pm$0.0)} & 0{\footnotesize ($\pm$0.0)} \\ \bottomrule
\end{tabular}
\end{table}

Finally, there is another aspect to consider to conclude this section of the study. Although all these results seem to point towards a great influence of CBT and destructive crossover side effects, one must not exclude the hypothesis that there is a matter of self-reinforcement (in this case, at a root node level). For that to be true, under our specific conditions, the dynamics involved in PIMP's Mate Choice mean that Choosers that prefer single-noded partners have a higher chance of producing better offspring. As such, if such a hypothesis held true, then in this case Subtree mutation would be the one acting against a natural process of self-reinforcement. 

In the original work, the authors provided a detailed analysis of the correlation between preferences (more specifically, attractiveness), and fitness, also evaluating the quality of the chosen mates fitness-wise, depicting a common trend of a negative correlation although with no clear direct linkage. Instead of comparing partners, we took a different approach: establishing a relation between preference depth and offspring quality. With this, we can evaluate whether there is evidence that convergence to a single node happens due to fitness gains or merely due to the lack of a selective force. To do this, a statistical correlation study between depth frequency and fitness gains might be tricky. That is because, while preferences converge, the solution chromosome is still undergoing fitness pressure and therefore improving the overall fitness of the population, thus an increase in frequency does not necessarily mean advantages in converging. 

\begin{figure}[ht]
  \centering
  \includegraphics[width=4in]{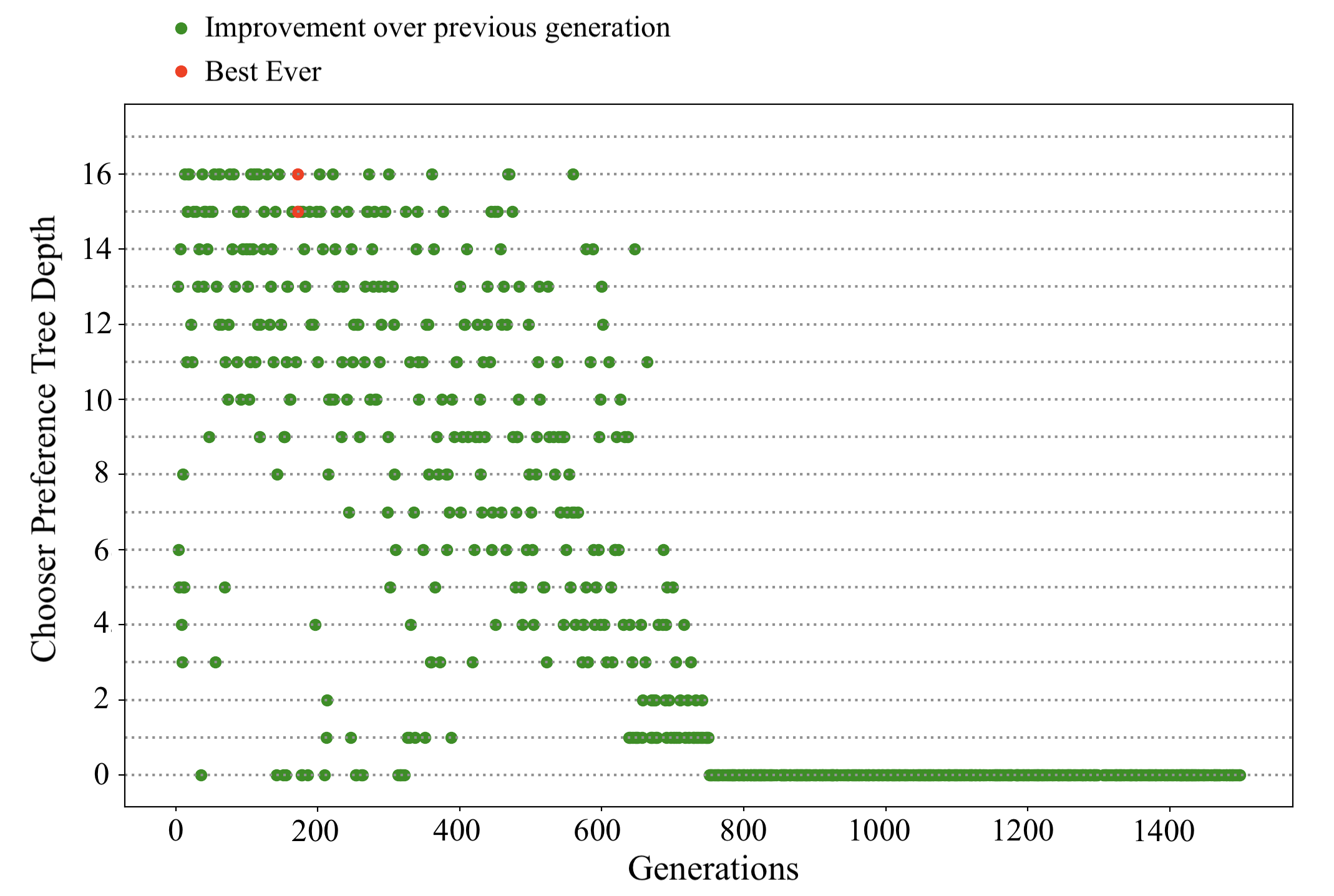}
  \caption{PIMP w/ Node Replacement mutation offspring fitness impact per depth -- Pagie-1.}
  \label{node_replace_leading_pagie1}
\end{figure}

\begin{figure}[h!]
  \centering
  \includegraphics[width=4in]{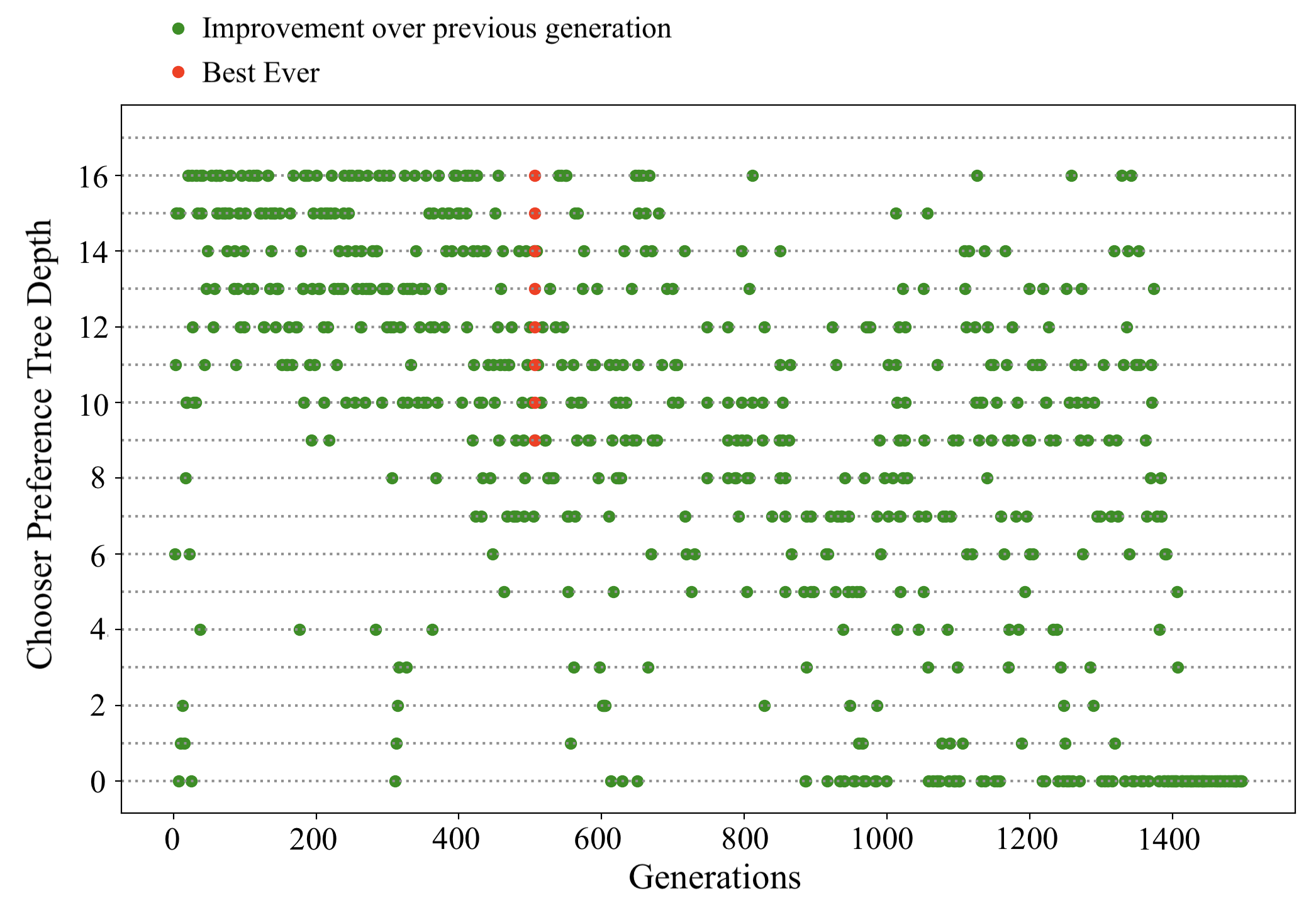}
  \caption{PIMP without mutation offspring fitness impact per depth -- Pagie-1.}
  \label{no_mut_leading_pagie1}
\end{figure}

As such, we tracked fitness improvements relative to the previous generation, sectioning each by preference depth of the first parent (the Chooser). More specifically, every time an offspring had better fitness than the best fitness of the previous generation, we marked the depth of the preference tree of the first parent.  Examples of the results are presented in Fig. \ref{node_replace_leading_pagie1} and \ref{no_mut_leading_pagie1}. 

\begin{figure}[h!]
  \centering
  \includegraphics[width=4in]{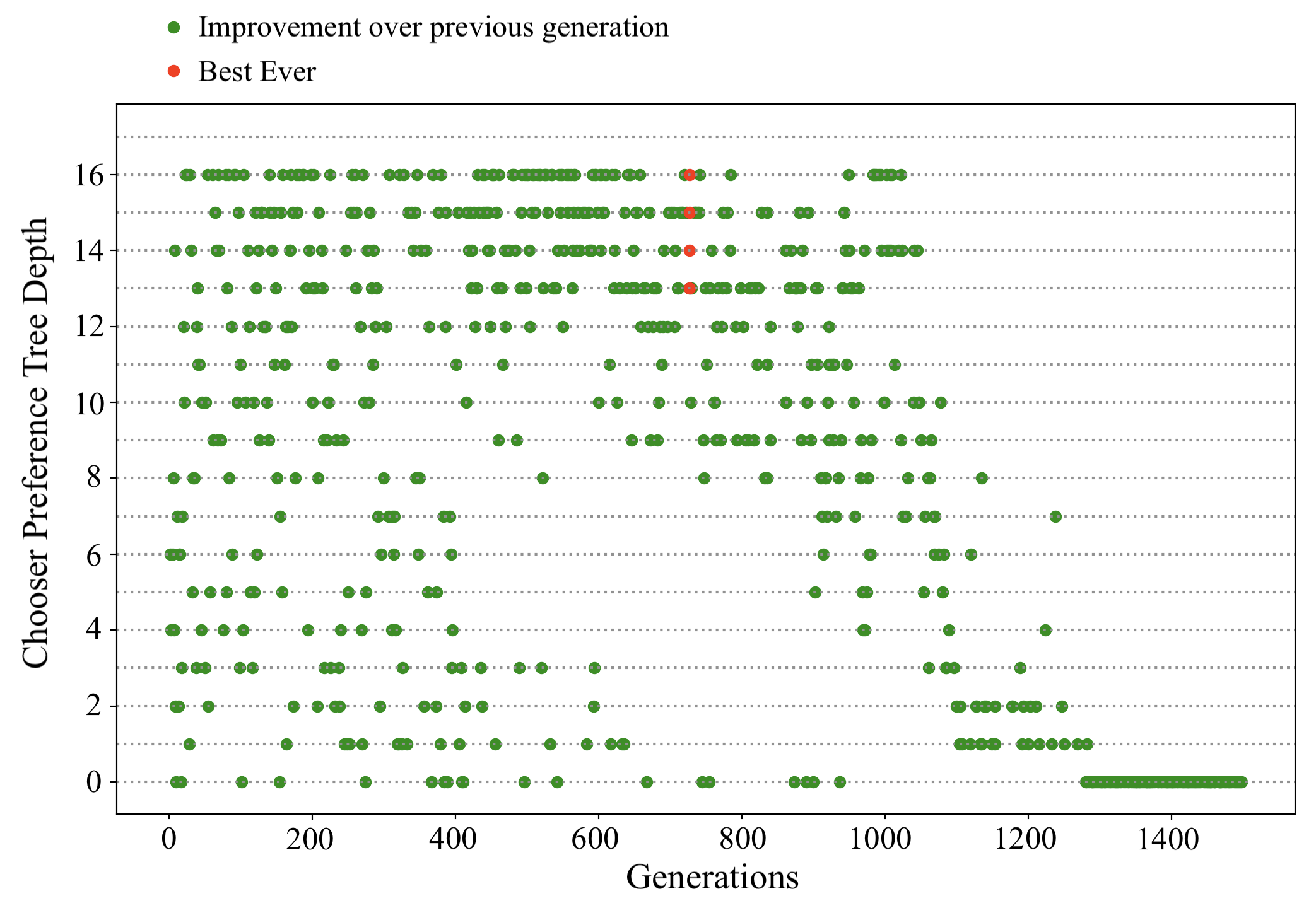}
  \caption{PIMP w/ hybrid mutation offspring fitness impact per depth -- Pagie-1.}
  \label{hybrid_leading_koza1}
\end{figure}

In all instances, preferences with a single node never reach the best solution ever (in red), and we can observe that preferences with a single node rarely provide the best child of the generation, only improving consistently after convergence. Fig. \ref{no_mut_leading_pagie1} is particularly a good example of this, where we have seen earlier that in this particular instance preferences converge slower to single-node (see Fig. \ref{subtree_depth_nomut_pagie1}), yet preferences with higher depth values also provide good offspring.
Finally, we include an instance of the hybrid-mutation version in Fig. \ref{hybrid_leading_koza1}, where we can again see that the convergence to a single node is unlikely to be caused by fitness.  Although improvements in fitness in relation to preference depth are quite even, preference trees with more depth (and thus potentially more complex) seem to provide the best offspring more regularly, but arguably this force is not enough to be maintained on its own as preference trees eventually evolve to a single unique node in the absence of Subtree mutation. This indicates that Mate Choice fails to stand on its own. 
Nevertheless, as shown in Table \ref{PIMP_subtree_avg_time_cost}, maintaining preferences comes at the cost of more running time. On average, the use of Subtree mutation almost doubles the time cost. From a theoretical standpoint, this increase is logical taking into account that larger preference trees take more time to be computed. In turn, with Node Replacementand or no mutation, preferences eventually converge to single node trees, which are much faster to compute. Interestingly, and despite this, these two last mutation types result in much higher variance (measured via Standard Deviation) than the Subtree mutation. 

\begin{table}[t!]
\caption{ PIMP Mutation Methdod Average Time Cost (in seconds).}\label{PIMP_subtree_avg_time_cost}
\renewcommand{\arraystretch}{1.5}
\begin{tabular}{cccc}
\toprule
\textit{} & Subtree & Node Replacement & No Mutation \\ \midrule
Koza-1 & 881.6{\footnotesize ($\pm$356.3)} & 410.7{\footnotesize ($\pm$291.0)} & 433.5{\footnotesize ($\pm$415.0)} \\
Nguyen-6 & 609.0{\footnotesize ($\pm$252.4)} & 221.6{\footnotesize ($\pm$210.0)} & 324.5{\footnotesize ($\pm$311.6)} \\ 
Pagie-1 & 951.4{\footnotesize ($\pm$418.3)} & 454.7{\footnotesize ($\pm$326.2)} & 465.0{\footnotesize ($\pm$462.3)} \\ 
\bottomrule
\end{tabular}
\end{table}
\vspace{2cm}

\subsection{Segregation and the role of each partner}
\begin{figure}[h!]
  \centering
  \begin{subfigure}{0.49\textwidth}
      \includegraphics[width=\textwidth]{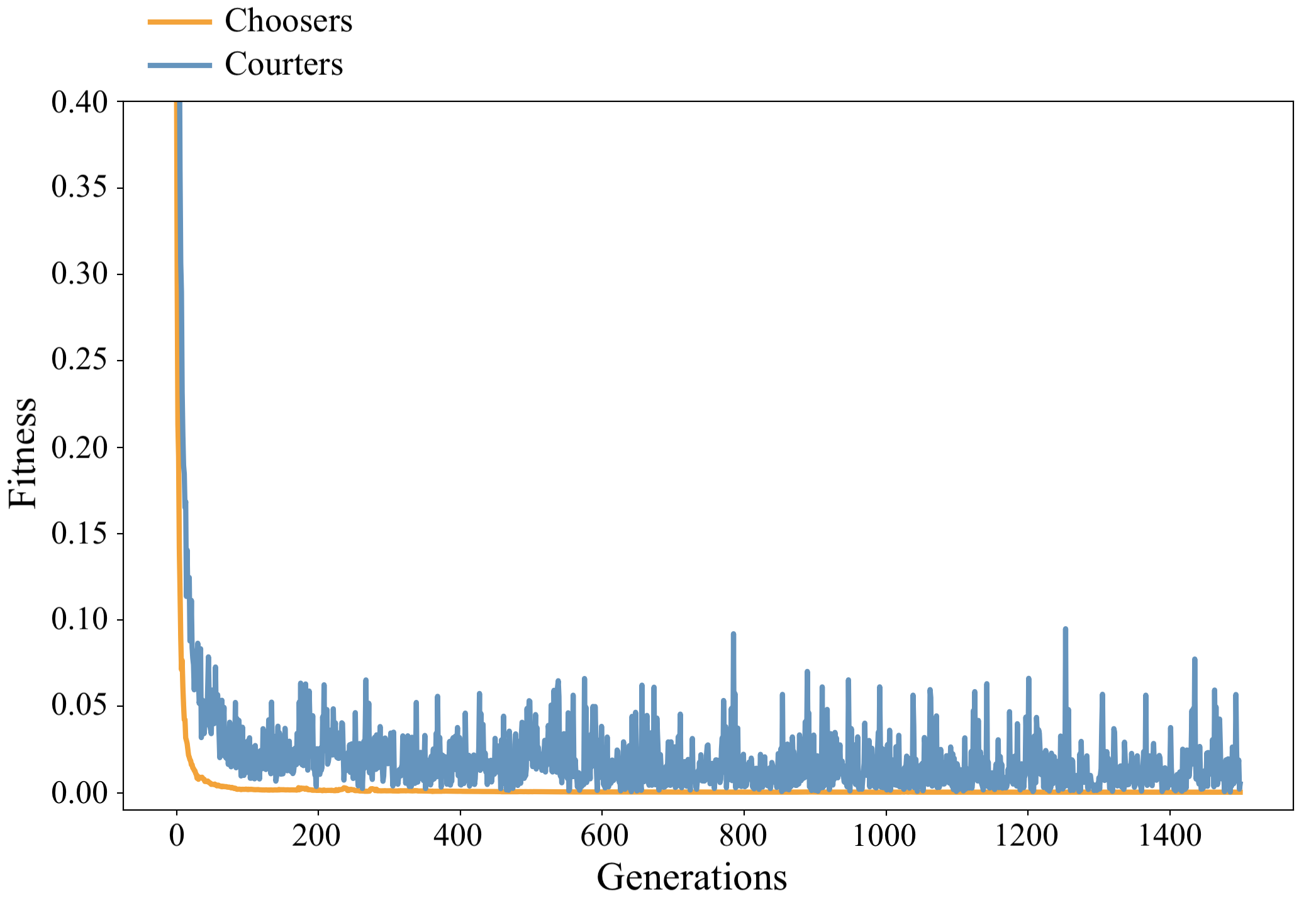}
      \caption{Subtree mutation.}
      \label{separated_mbf_subtree_koza1}
  \end{subfigure}
  \hfill
  \begin{subfigure}{0.49\textwidth}
      \includegraphics[width=\textwidth]{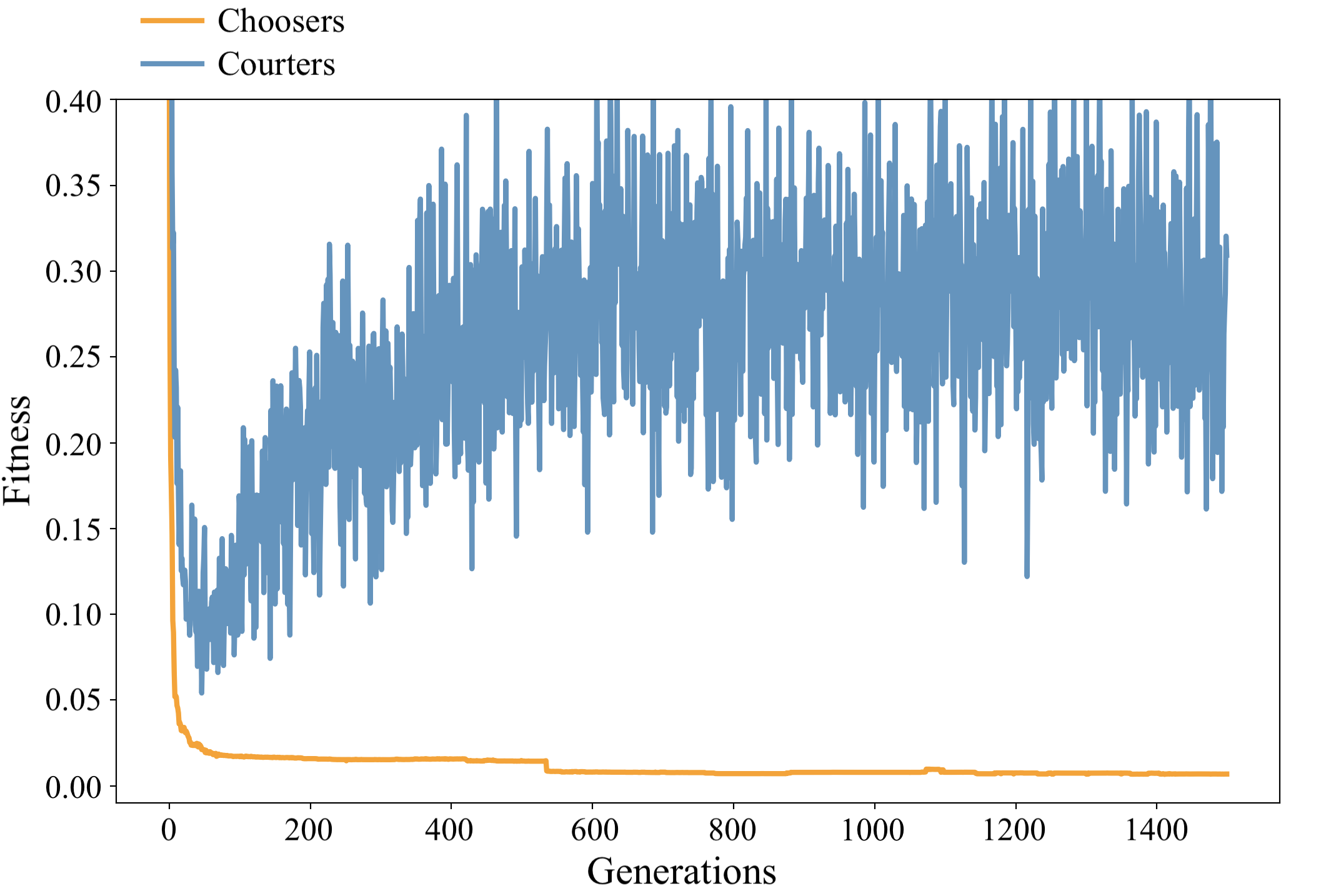}
      \caption{Node Replacement.}
      \label{separated_mbf_nodereplace_koza1}
  \end{subfigure}
  \begin{subfigure}{0.49\textwidth}
      \includegraphics[width=\textwidth]{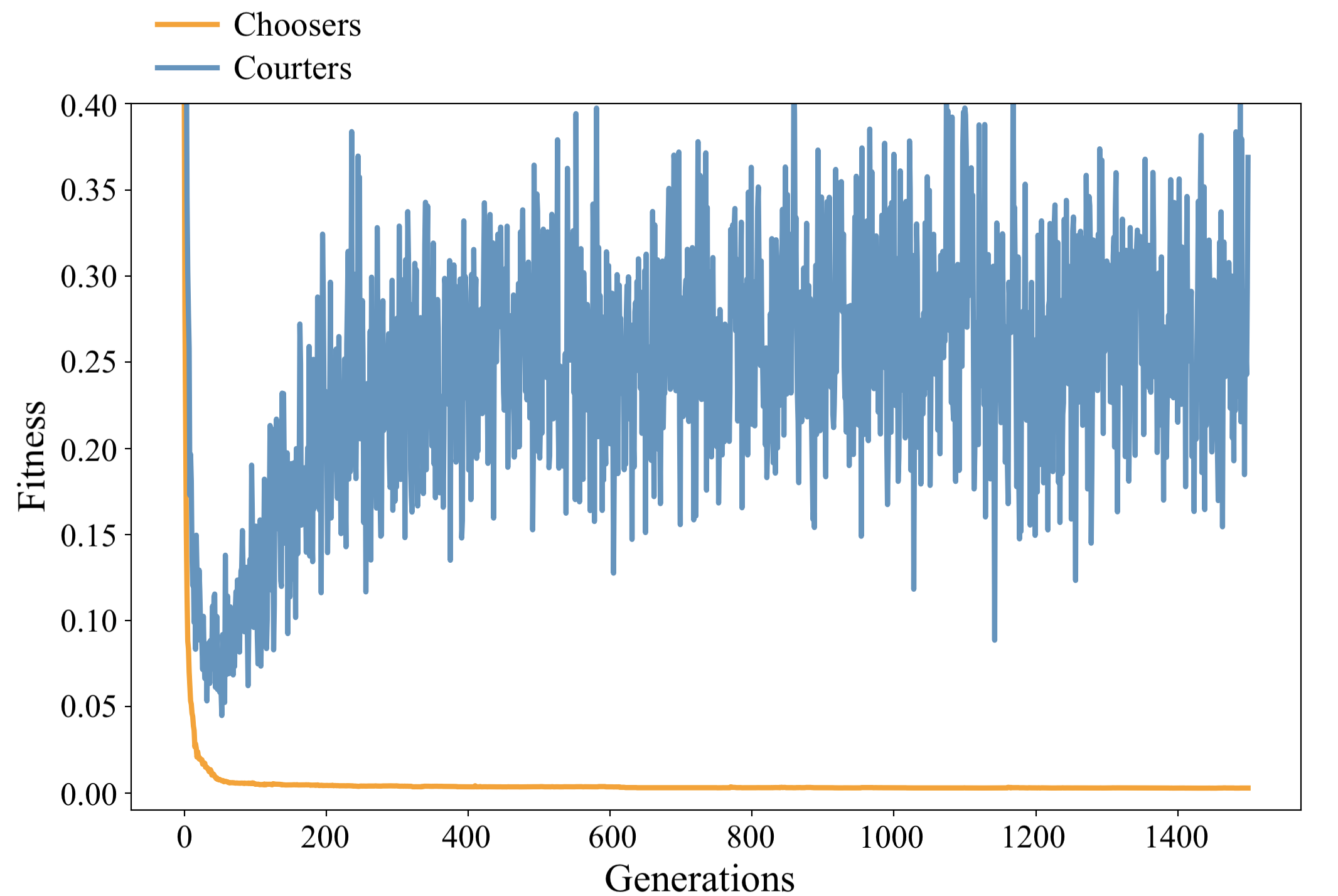}
      \caption{No Mutation.}
      \label{separated_mbf_nomut_koza1}
  \end{subfigure}
  \caption{Impact of mutation in MBF of both roles -- Koza-1.}
\end{figure}

As two different forces act upon Choosers and Courters, there is the potential of having contrasting evolutionary directions which is, in fact, the staple of PIMP. As such, the population tends to evolve into two distinct groups of specialised individuals. This was shown in a recent study \citep{SimoesEvostar}, similar to what was observed in the original work where the population quickly evolved into two different groups in a role segregation fashion, resembling the phenomena often observed powered by Sexual Selection. The study has also pointed out that as the evolutionary process carries on, the more the roles diverge when comparing partners, although instances of assortative mating were found. Furthermore, it was also stated that while Choosers assume a more conservative and exploitative nature (most likely due to fitness pressure), Courters appear to follow a more explorative one. This characteristic of the Courters can mainly be seen through a fitness-oriented view, as briefly addressed in \citep{SimoesEvostar} and now formally demonstrated in Fig. \ref{separated_mbf_subtree_koza1}, \ref{separated_mbf_nodereplace_koza1} and \ref{separated_mbf_nomut_koza1}. We must clarify that, for the sake of readability, individuals that acted as both roles (on average between 5\% and 10\% of the population) were left out of this analysis, as their overall Mean Best Fitness was similar to that of Courters.

These figures illustrate how the MBF of each role evolves, where we can observe that Choosers assume the best fitness path of the population, while Courters tend to diverge and evolve to progressively worse fitness landscapes. The results also reveal that preference convergence (in the absence of subtree mutation) tends to promote less fit Courters when compared to sustainable preference trees (Fig. \ref{separated_mbf_subtree_koza1}).

Again, looking at depth as a structure feature, results also point towards the same direction: while Choosers grow, on average, to the highest depth possible (often observed in GP due to bloat \citep{Poli2007, Trujillo2019, Luke2006, Trujillo2016}), Courters tend to assume smaller sizes on average, thus potentially less complex trees. 

\begin{figure}[t]
 \centering
 \begin{subfigure}{0.49\textwidth}
     \includegraphics[width=\textwidth]{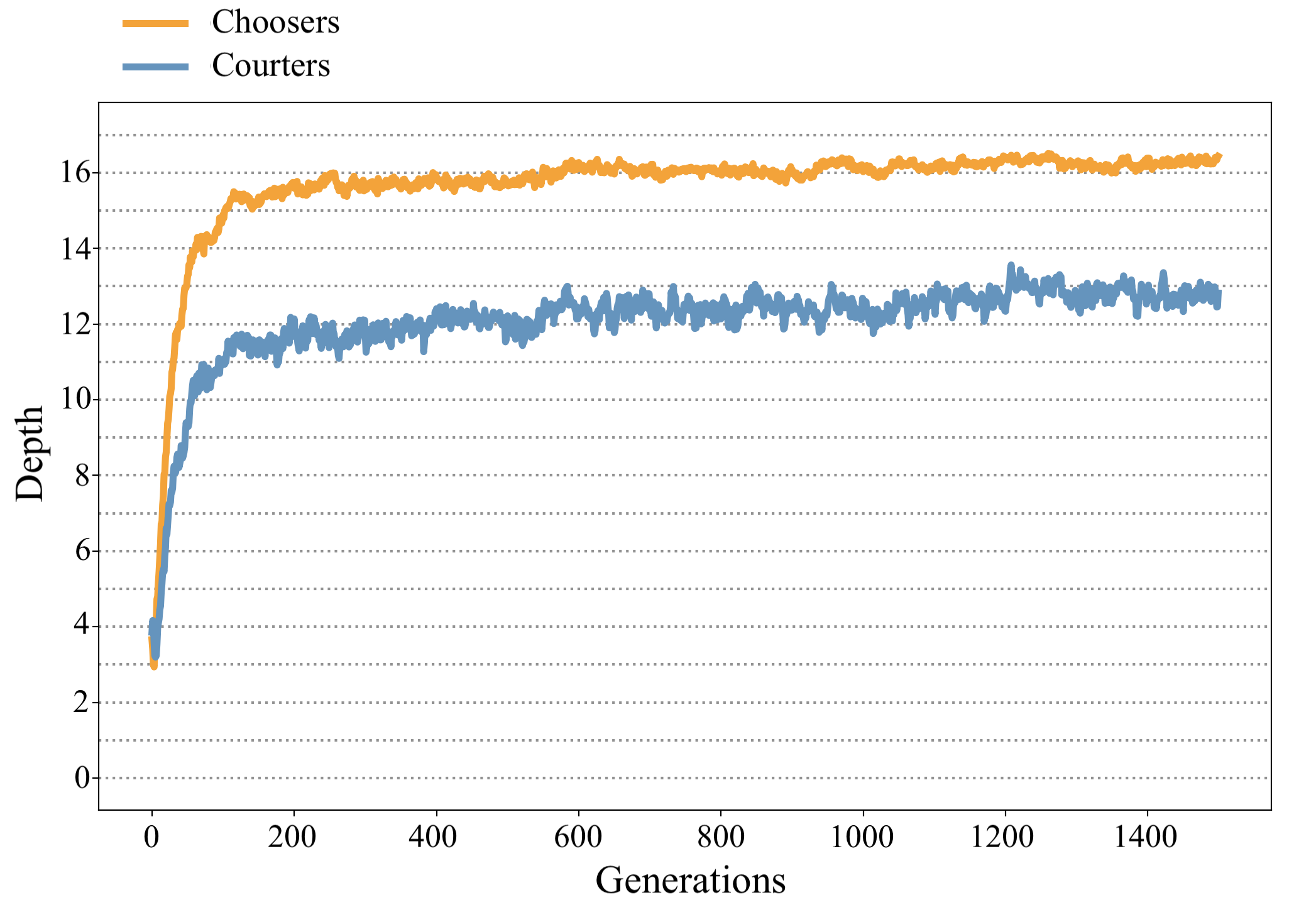}
     \caption{Subtree Mutation.}
     \label{sep_depth_subtree_koza1}
 \end{subfigure}
 \hfill
 \begin{subfigure}{0.49\textwidth}
     \includegraphics[width=\textwidth]{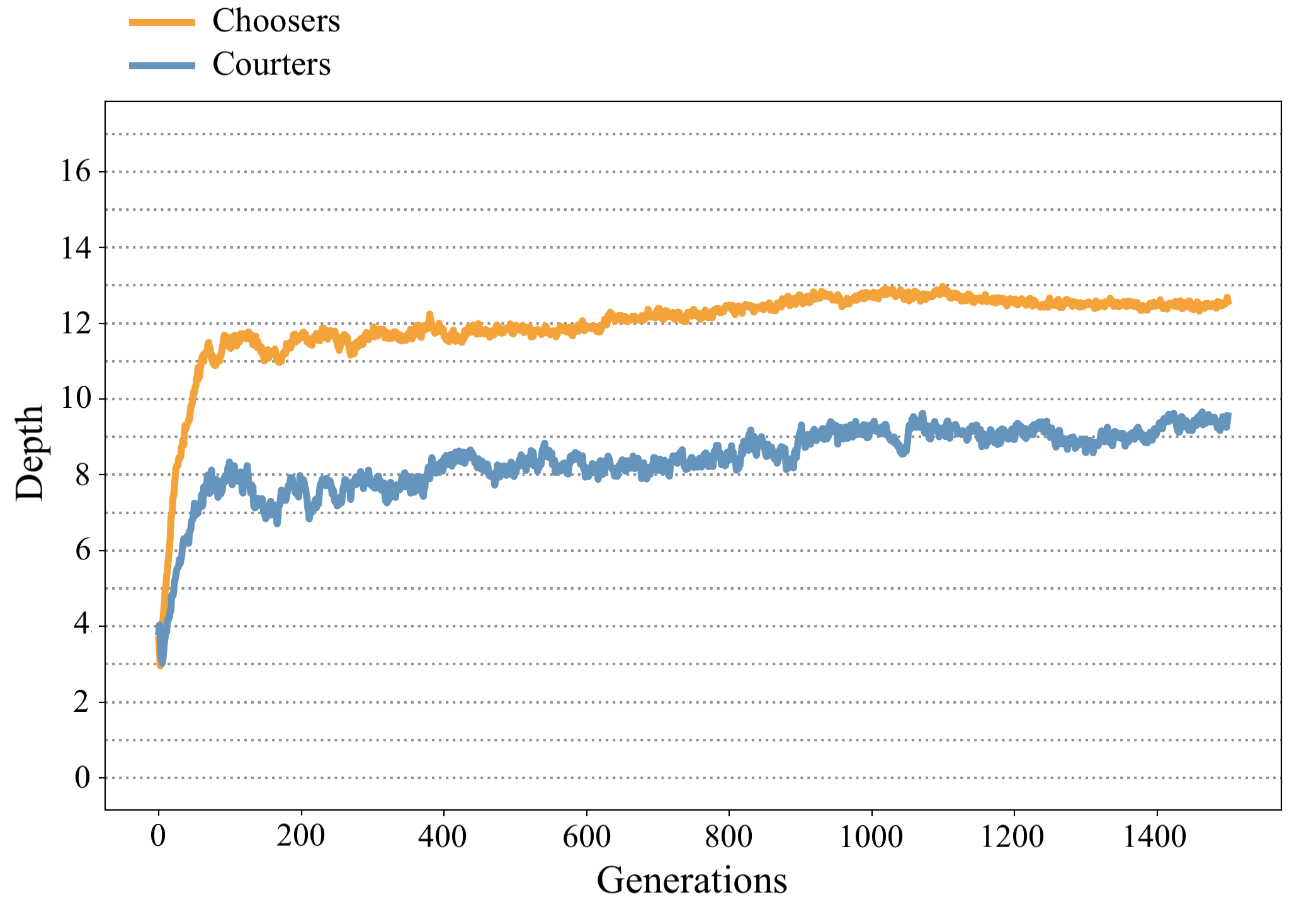}
     \caption{No Mutation.}
     \label{sep_depth_nomut_koza1}
 \end{subfigure}
 \caption{PIMP Average Roles Depth -- Koza-1.}
\end{figure}

Furthermore, one interesting aspect of this is that while the global fitness of the Courters is poor, they actually preserve smaller trees within the population. As examples instances, Fig. \ref{sep_depth_subtree_koza1} and \ref{sep_depth_nomut_koza1} demonstrate the evolution of the average depth of the roles.

\begin{figure}[!h]
\centering
\begin{subfigure}{0.9\textwidth}
	\includegraphics[width=\textwidth]{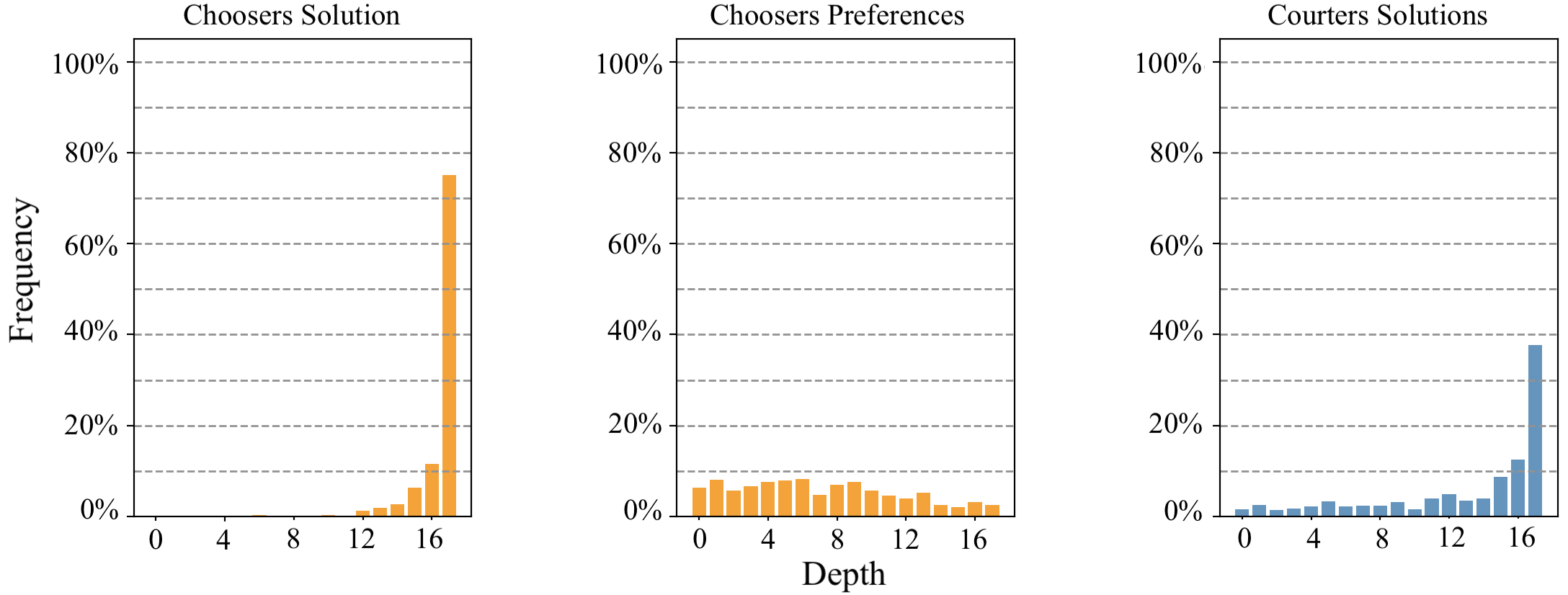}
	\caption{PIMP w/ Subtree Mutation.}
	\label{PIMP_Subt_Depth}
\end{subfigure}
\hfill
\begin{subfigure}{0.9\textwidth}
	\includegraphics[width=\textwidth]{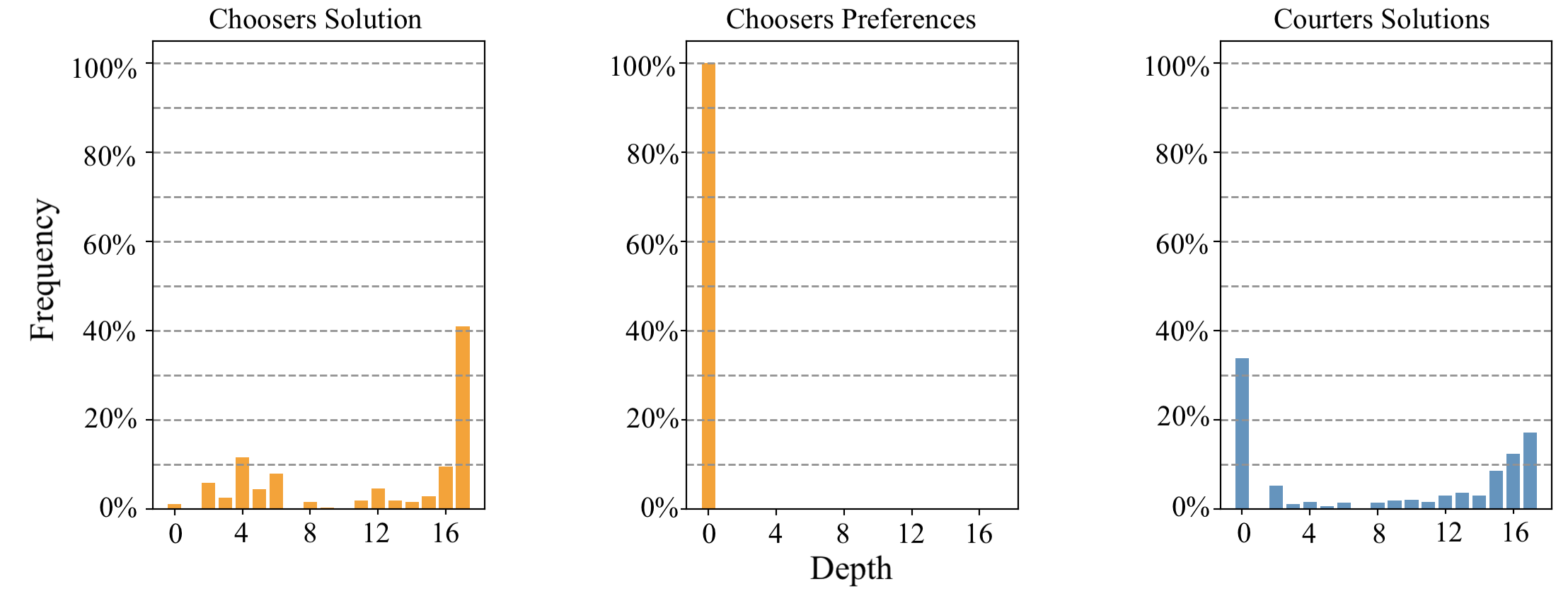}
	\caption{PIMP without Mutation.}
	\label{PIMP_Not_Mut_Dpeth}
\end{subfigure}
\caption{Average Depth Frequency Distribution per role at termination -- Koza-1.}
\end{figure}

Finally, all these variations in fitness and overall depth led us to question whether these different pressures had any impact on the depth distribution of each role. Considering that preferences are able to evolve freely (at least until falling into single-node convergence), it would be interesting to know if Courters tend to converge alongside preferences and the impact it has on Choosers. For that, we measured the frequency of the tree depth of Courters and Choosers at termination of each run. As Fig.  \ref{PIMP_Subt_Depth} demonstrates, the differences in average solution depth described earlier are also visible here, yet we observe that while Choosers in fact tend to evolve to larger solutions Courters seem to evolve to be a set of more varied tree structures. When comparing this case against the same target function but eliminating mutation, we observed, just as discussed earlier, that preferences converge to the simplest possible form. Yet, by studying the depth distribution separately, we can see that this force not only affects the Courters, which evolved towards simpler solutions, but it also affected Choosers, most likely as a side effect of preferences itself: by preferring smaller trees, Courters are more likely to evolve towards single-node tress, which in turn means that Choosers will be recombined with smaller trees and thus are less likely to grow. 
 
Note that although only one test instance is shown for the sake of parsimony, the behaviour was similar in all testing instances.

\subsection{Comparing with Tournament Selection}
To conclude our analysis, we believe that it is important to establish a direct comparison between these different set-ups and a Tournament selection. For that, we used a standard tournament selection of size = 5 with the same setup as described in Table \ref{Set_up}). It was shown in the original work that PIMP generally promoted an improvement over a standard approach, yet in this experiment we are not only extending the study to three mutation methods but also letting the evolution proceed for longer generations. We extend this comparison by using two metrics beyond fitness (i.e., MBF): Solution Diversity (number of unique trees), and Tree Depth. In addition to the three functions explored in the former sections, we also performed tests on the Diabetes dataset available at the SciPy library \citep{2020SciPy}, which comprises 442 samples from diabetes patients, each holding ten different features (i.e., baseline variables) and a quantitative measure of disease progression (i.e., target variable). 
This allows us to compare PIMP to a standard tournament selection under a considerably more challenging real-world problem. Furthermore, minor changes were made in this regard. First, the function set was updated to include constants and more operations (see Table \ref{diabetes_function_set}), and only subtree mutation was used. Secondly, termination conditions were shortened to 500 generations to make for the required expensive computation.  As such, results gathered from this dataset will be presented separately to facilitate readability.

\begin{table}[bp]
	\caption{Function Set for the Diabetes dataset.}\label{diabetes_function_set}
	\renewcommand{\arraystretch}{1.5}
	\begin{tabular}{cc}
		\hline
		Functions & Constants \\ \hline
		\multicolumn{1}{c}{$+$, $-$, $\times$, $\%$, $\sin$, $\cos$, $e^n$, $\ln(|n|)$, $\sqrt{n}$, $|n|$}  & \multicolumn{1}{c}{integer(-10,10)} \\ \hline
	\end{tabular}
\end{table}

Statistical tests were performed to assess the impact of the observed differences. Shapiro-Wilk tests were performed to assess the likelihood of having normally distributed data, along with Bartlett’s tests to assess the variance between groups. As mostly all data was unlikely to follow a Gaussian distribution for an alpha of 5\% (only two pairs of instances pointed towards a likelihood of a normal distribution, yet showed no signs of equal variance, also for an alpha of 5\%), we performed the non-parametric Wilcoxon signed rank test between pairs. For the following tables, boldface is used to identify the best results, and an asterisk symbol (\textbf{*}) is used to flag instances where statistically significant differences were found between the PIMP and tournament selection versions being tested.

As presented in Tables \ref{PIMP_vs_Std_MBF} and \ref{Diabetes_Results}, results are mainly competitive, with no clear signs of advantage in using one approach over the other. Statistically significant differences were found in Nguyen-6 (subtree mutation and no mutation) where PIMP performed better, while in Koza-1 with Node Replacement and the diabetes dataset the standard tournament selection was able to surpass PIMP.

\begin{table}[]
\caption{MBF Comparision - PIMP vs. Tournament Selection.}\label{PIMP_vs_Std_MBF}
\renewcommand{\arraystretch}{1.2}
\tabcolsep12pt
\begin{tabular}{lcc}
\toprule
& PIMP & Tournament Selection \\ \cmidrule(lr){2-3}
& \multicolumn{2}{c}{Subtree Mutation} \\ \cmidrule(lr){2-3}
Koza-1   & \textbf{3.7E-4}  ($\pm$ 6.6E-4) & 1.4E-3  ($\pm$ 2.5E-3)  \\ 
Nguyen-6 & \textbf{6.2E-4*} ($\pm$ 1.8E-3) & 3.3E-3* ($\pm$ 1.0E-2)  \\ 
Pagie-1  & 1.5E-2  ($\pm$ 1.2E-2) & \textbf{1.0E-2 } ($\pm$ 1.1E-2)  \\ \cmidrule(lr){2-3}
& \multicolumn{2}{c}{Node Replacement Mutation} \\ \cmidrule(lr){2-3}
Koza-1   & 6.9E-3* ($\pm$ 2.1E-2) & \textbf{5.9E-4*} ($\pm$ 1.6E-3) \\
Nguyen-6 & 7.7E-3  ($\pm$ 2.3E-2) & \textbf{3.6E-3}  ($\pm$ 1.4E-2) \\
Pagie-1  & 4.3E-2 ( $\pm$ 1.6E-1) & \textbf{1.9E-2}  ($\pm$ 1.5E-2) \\ \cmidrule(lr){2-3}
& \multicolumn{2}{c}{No Mutation} \\ \cmidrule(lr){2-3}
Koza-1   & 2.8E-3  ($\pm$ 1.1E-2) & \textbf{1.3E-3}  ($\pm$ 3.7E-3) \\
Nguyen-6 & \textbf{5.7E-3*} ($\pm$ 1.6E-2) & 8.4E-3* ($\pm$ 2.1E-2) \\
Pagie-1  & \textbf{1.8E-2}  ($\pm$ 2.1E-2) & 2.4E-2  ($\pm$ 1.9E-2) \\
\bottomrule
\end{tabular}
\end{table}

\begin{table}[b]
\caption{Average Unique Solutions at Termination -- PIMP vs. Tournament Selection.}\label{PIMP_vs_Std_Unique_Trees}
\renewcommand{\arraystretch}{1.2}
\tabcolsep12pt
\begin{tabular}{lcc}
\toprule
& PIMP & Tournament Selection \\ \cmidrule(lr){2-3}
& \multicolumn{2}{c}{Subtree Mutation} \\ \cmidrule(lr){2-3}
Koza-1   & \textbf{86\%*} ($\pm$6) & 62\%*($\pm$14)  \\ 
Nguyen-6 & \textbf{84\%*} ($\pm$5) & 57\%*($\pm$12)  \\ 
Pagie-1  & \textbf{90\%*} ($\pm$5) & 66\%*($\pm$16)  \\ \cmidrule(lr){2-3}
& \multicolumn{2}{c}{Node Replacement Mutation} \\ \cmidrule(lr){2-3}
Koza-1   & \textbf{60\%}  ($\pm$33)   & 58\%  ($\pm$32) \\
Nguyen-6 & \textbf{57\%}  ($\pm$35)   & 49\%  ($\pm$16) \\
Pagie-1  & \textbf{83\%*} ($\pm$18) & 59\%* ($\pm$15) \\ \cmidrule(lr){2-3}
& \multicolumn{2}{c}{No Mutation} \\ \cmidrule(lr){2-3}
Koza-1   & \textbf{58\%}  ($\pm$32) & 44\%  ($\pm$17) \\
Nguyen-6 & \textbf{54\%}  ($\pm$34) & 45\%  ($\pm$15) \\
Pagie-1  & \textbf{81\%*} ($\pm$17)   & 56\%* ($\pm$20) \\
\bottomrule
\end{tabular}
\end{table}

In contrast, when it comes to unique tree diversity (Tables \ref{PIMP_vs_Std_Unique_Trees} and \ref{Diabetes_Results} ), PIMP seems to offer a solid alternative to a standard tournament selection, consistently providing more different solutions. Once again, Subtree mutation seems to give PIMP a larger advantage as it always resulted in significant statistical differences. The same goes for the tournament selection where Subtree mutation naturally promoted more diverse trees, yet these results suggest that by having a Mate Choice mechanism such as in PIMP the effects regarding tree diversity are even larger. Furthermore, we observe that under Subtree mutation PIMP got lower standard deviation scores than the tournament selection, particularly regarding fitness and percentage of unique trees. These results reinforce the notion that PIMP is quite sensitive to the mutation method used, as the standard deviation scores show that variation increases drastically in the absence of Subtree mutation.  On a side note, we observe that in all Pagie-1 instances PIMP holds higher diversity values even in the absence of mutation. This is likely due to the fact that this particular function has two variables, meaning that even when preferences converge to single nodes -- as shown earlier -- it can assume two different values nonetheless. Thus, although preferences converge to the smallest depth, it is still possible for two different preferences to be maintained. Results from the diabetes dataset (Table \ref{Diabetes_Results}) indicate the same tendency, with PIMP promoting more diverse trees.

\begin{table}[]
\caption{Average Tree Depth at termination - PIMP vs. Tournament Selection.}\label{PIMP_vs_Std_Depth}
\renewcommand{\arraystretch}{1.2}
\tabcolsep12pt
\begin{tabular}{lcc}
\toprule
& PIMP & Tournament Selection \\ \cmidrule(lr){2-3}
& \multicolumn{2}{c}{Subtree Mutation} \\ \cmidrule(lr){2-3}
Koza-1   & \textbf{14.2*} ($\pm$1.3) & 16.5* ($\pm$0.6)  \\ 
Nguyen-6 & \textbf{14.1*} ($\pm$1.0) & 16.0* ($\pm$2.1)  \\ 
Pagie-1  & \textbf{14.2*} ($\pm$1.2) & 16.4* ($\pm$1.0)  \\ \cmidrule(lr){2-3}
& \multicolumn{2}{c}{Node Replacement Mutation} \\ \cmidrule(lr){2-3}
Koza-1   & \textbf{10.1*} ($\pm$5.5) & 16.1* ($\pm$0.9) \\
Nguyen-6 & \textbf{10.0*} ($\pm$5.9) & 13.9* ($\pm$4.0) \\
Pagie-1  & \textbf{13.3*} ($\pm$3.2) & 15.5* ($\pm$1.5) \\ \cmidrule(lr){2-3}
& \multicolumn{2}{c}{No Mutation} \\ \cmidrule(lr){2-3}
Koza-1   & \textbf{10.7*} ($\pm$5.6) & 13.5* ($\pm$4.5) \\
Nguyen-6 & \textbf{10.6*} ($\pm$5.7) & 14.2* ($\pm$4.2) \\
Pagie-1  & \textbf{13.5}  ($\pm$2.7) & 15.0  ($\pm$1.9) \\
\bottomrule
\end{tabular}
\end{table}

\begin{table}[h!]
\caption{Diabetes dataset results (at termination).}\label{Diabetes_Results}
\renewcommand{\arraystretch}{1.5}
\begin{tabular}{ccc}
\toprule
 & PIMP & Tournament Selection \\ \midrule
MBF & 3.1E-3*{\footnotesize ($\pm$2.9E2)} & \textbf{2.8E-3*}{\footnotesize ($\pm$3.1E2)} \\
Avg Unique Trees & \textbf{94\%*}{\footnotesize ($\pm$3.4)} & 72\%*{\footnotesize ($\pm$8.9)} \\
Avg Tree Depth & \textbf{13.5*}{\footnotesize ($\pm$1.1)} & 16.5*{\footnotesize ($\pm$0.8)} \\ \bottomrule
\end{tabular}
\end{table}

Finally, PIMP consistently promotes smaller solutions on average, arguably due to the force of preferences that are generally smaller, as we can observe in Tables \ref{PIMP_vs_Std_Depth} and \ref{Diabetes_Results}. As a result, Courters tend to develop smaller structures than Choosers, and since they roughly make up half the population, it also affects the average solution depth of the population. In this experiment, these are critical benefits of using PIMP, as statistical differences were found in all instances except for one (No Mutation), showing that PIMP consistently promotes smaller trees than a standard tournament selection. This might also contribute to a more diverse depth distribution that is arguably quite difficult to achieve with a tournament selection alone, mostly due to the constant fitness pressure acting upon both parents (with bloat also contributing to it). 

\begin{figure}[h!]
 \centering
 \begin{subfigure}{0.6\textwidth}
     \includegraphics[width=\textwidth]{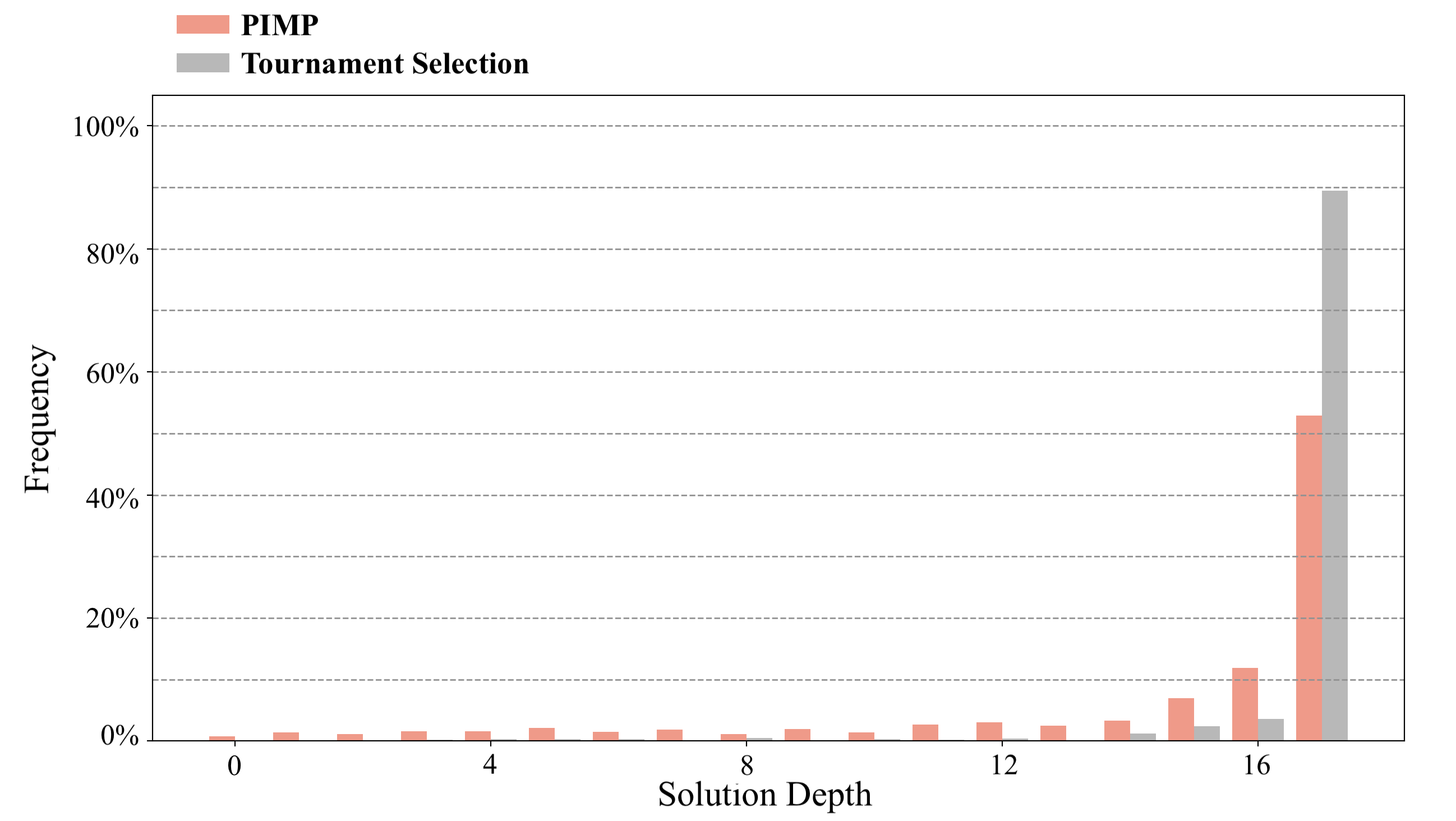}
     \caption{Subtree Mutation (Koza-1).}
     \label{koza1_subtree_disti}
 \end{subfigure}
 \begin{subfigure}{0.58\textwidth}
     \includegraphics[width=\textwidth]{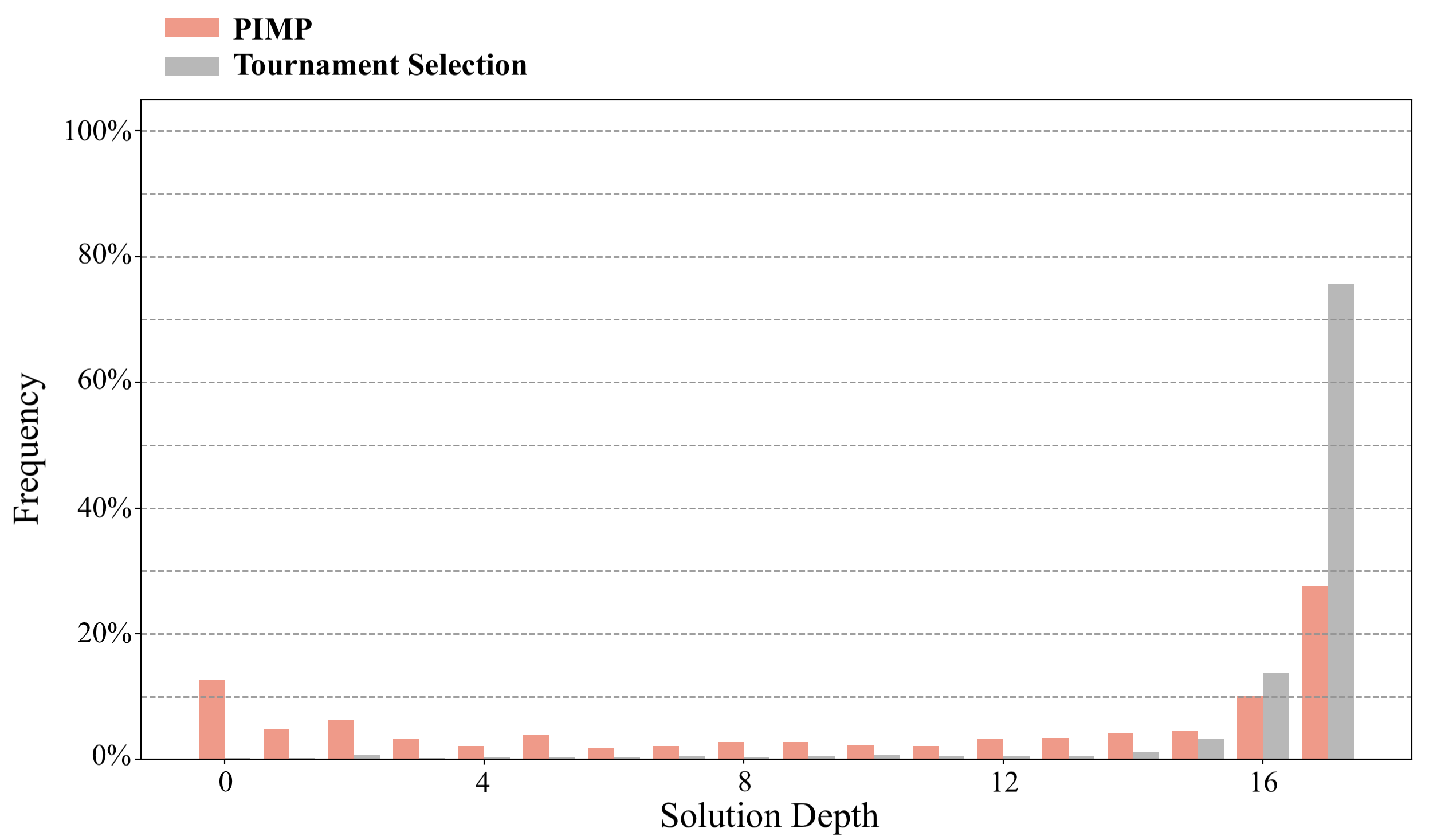}
     \caption{Node Replacement (Koza-1).}
     \label{koza1_NodeReplace_disti}
 \end{subfigure}
 
 \begin{subfigure}{0.64\textwidth}
     \includegraphics[width=1\textwidth]{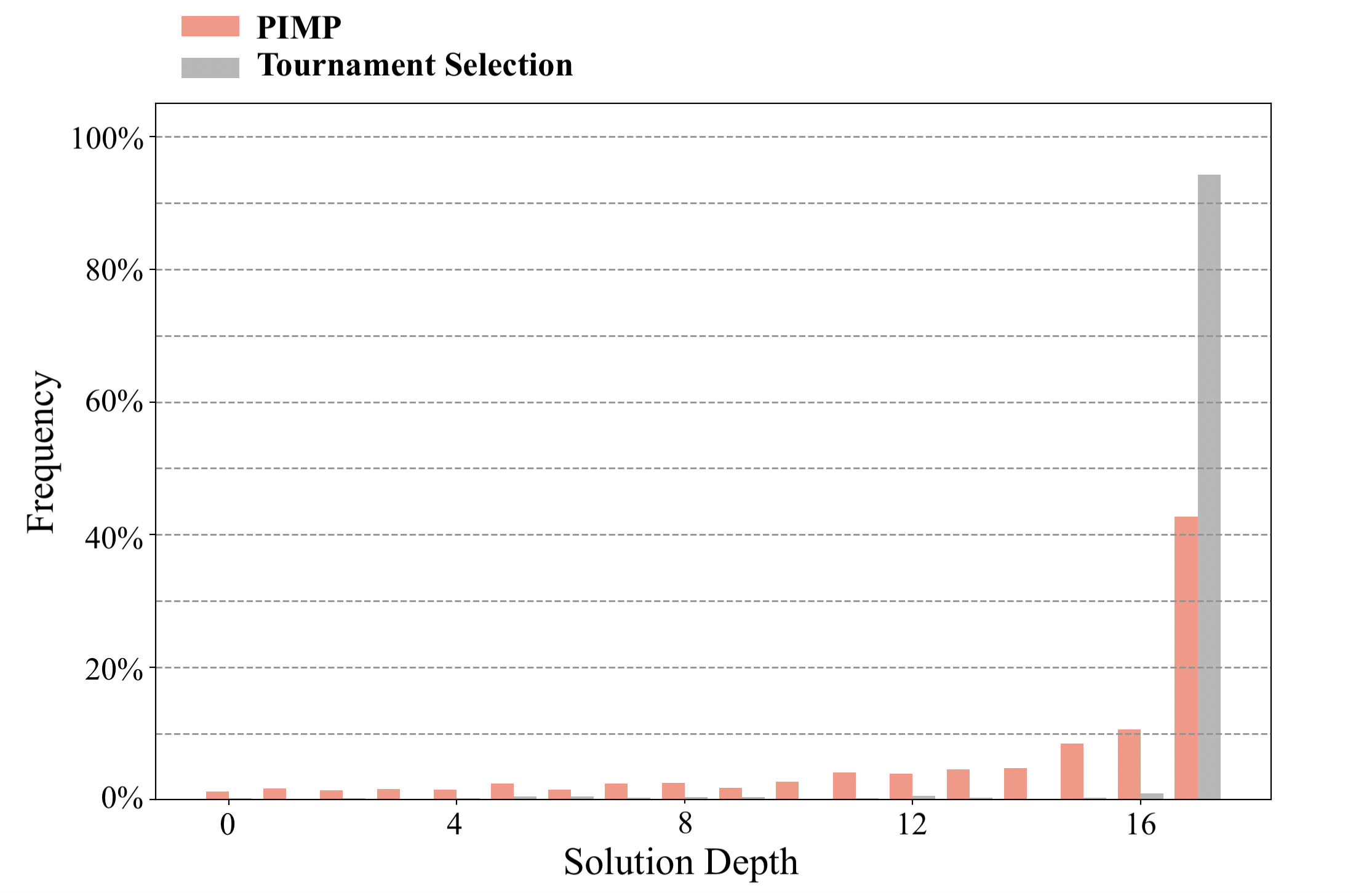}
     \caption{Subtree Mutation (Diabetes).}
     \label{DiabetesDepth}
 \end{subfigure}
 \caption{Solution Depth Distribution at Termination.}
\end{figure}

We can observe this distinction with two examples that illustrate precisely this, as seen in Fig. \ref{koza1_subtree_disti}, \ref{koza1_NodeReplace_disti} and \ref{DiabetesDepth}, with data from the last generation of all runs. We notice that with subtree mutation, under a standard tournament selection most of the individuals have the highest depth value, while PIMP attenuates this concentration. Also, when Node Replacement mutation is used (and preferences converge) PIMP offers a more balanced depth distribution, mostly because all Choosers prefer single-noded trees. 

\section{Discussion}
The results addressed in the previous section allowed us to better understand what is actually happening in the evolution of an ideal mate representation. As the original work provides most of its analysis over the dynamics of PIMP with no mutation at all, we have found evidence that the diversity promoted by the method is unlikely to be caused by a natural evolution towards direct fitness benefits. Nonetheless, even if preferences fail to create some sort of self-reinforcement dynamics and guide Mate Choice, diversity is maintained (as we have seen, when preferences converge to single nodes it seems to be the result of an absence of selective pressure rather than by self-reinforcement alone).

Arguably, one of the most interesting notions that we can extract from these results lies precisely on this point. While the solution chromosomes of the Choosers evolve to maximum depth (theoretically due to bloat), preference chromosomes suffer from exactly the opposite effect by having no fitness pressure: it naturally falls to the simplest solution possible, something likely to be explained by the Crossover bias theory. In essence, when mutation is excluded, diversity is promoted in PIMP not necessarily by coevolution of preferences but rather by two contrasting phenomena working in tandem: one pushes the population to higher depths, while the other keeps part of it to more evenly distributed trees (often pushing solutions towards smaller sizes). Although our results point towards this explanation, further tests are required to support this possibility.

Regarding the method itself, our results show that a Subtree mutation is more suitable not only in terms of performance but also when it comes to maintaining solution diversity, at least when facing symbolic regression problems. Nonetheless, in all mutation variations used in this experiment, PIMP always promoted more unique trees than a standard tournament selection, although differences were not as drastic when mutation without affecting tree depth was introduced. However, when the target function had two variables (i.e., Pagie-1), PIMP was still able to maintain high levels of diversity regardless of the mutation type being used, which suggests that PIMP has nonetheless the potential of boosting diversity whenever preferences do not converge. As such, our results indicate that the method suffers from a secondary phenomenon observed in GP and, although this does not compromise its performance drastically, it seems to benefit largely from having a healthy preference distribution when it comes to diversity. 

Finally, and in line with the previous note, we must stress that Mate Choice is quite an intricate phenomenon to study, and while our work points towards a possible limitation of the method (when it comes to a natural preference self-reinforcement), it still manages to maintain diversity. Moreover, it is also worth noting that when preferences are sustained by subtree mutation it is still unclear if they are able to evolve naturally towards a stability point. This topic can serve as a starting point for future research. It should also be noted that although our results are consistent, they are hardly generalizable as more data should be gathered, for example on a larger set of symbolic regression instances while targeting a single performance or diversity metric. In the future, it would also be interesting to expand the benefits of using PIMP by comparing its performance against other proposed methods for diversity maintenance in GP, and even studying how different preference structures affect the overall performance (e.g., binary strings instead of GP trees).
In addition, the impact mating preferences have on the evolutionary process opens possibilities for various approaches to explore. Specifically, the use of mating preferences could be tailored to prioritize smaller solutions if the fitness gains are comparable. 
It might also be useful in other fields beyond symbolic regression. For instance, the diversity promoted by mating preferences can be valuable when addressing multiple target points or objectives. A diverse set of preferences can target different points of interest, which may be useful when dealing with weighted training points or multiple objectives.

\section{Conclusion}
In nature, Sexual Selection is known to work alongside Natural Selection promoting physical and behavioural differences between sexes in search for a mate. The idea of having another force beyond fitness pressure has been explored in the past as a way to promote and maintain diversity within populations in EAs in general, but also in GP. Particularly at the parent selection stage, most frequently we observe methods promoting dissimilar or similar couples as a way to avoid early convergence. Although these methods resemble, to some extent, a general outcome of Sexual Selection, some of its core notions are often left unexplored, likely due to its complexity and intricate dynamics. The PIMP method has been proposed recently and tackles precisely one of these premises by modelling Mate Choice through ideal mate representations working as mating preferences in GP. 

In this article, we extended the knowledge on the method, experimenting with different mutation variations while studying how mating preferences evolve under three symbolic regression instances. Results show that the evolution of preferences is highly affected by the type of mutation used. More precisely, we found that, in the absence of a mutation that widens tree depth, preferences tend to evolve to the simplest structure possible, eventually converging into single-node trees. Furthermore, a closer look at local fitness demonstrates that simpler preferences do not necessarily result in better offspring, indicating that this convergence is likely to be occurring due to the lack of selective pressure rather than natural self-reinforcement. Nonetheless, this lack of diversity in preferences still promotes solution diversity by establishing two opposing forces: while the first parent is subject to fitness, partners are chosen based on the first parent's preferences. If we consider that single-node trees are not similar to the target function, then roughly half the population is pursuing a different goal. In fact, this still promotes a clear role segregation, with Courters assuming a more diverse depth distribution than Choosers who evolve to be the fittest individuals.  

We finally compared different PIMP versions to a standard tournament selection also with different mutation types and a real-world dataset, and while PIMP did not always outperform the latter (measured through mean best fitness), it did promote more diversity regarding unique solutions, reaching statistical significance in all instances where Subtree mutation was used. Moreover, it was also found that PIMP tends to promote smaller solutions in general, failing to reach statistical significance in only one out of the ten comparisons made. Furthermore, we found a clear tendency of mating preferences to act against the known phenomenon of bloat (at least when coupled with a depth restriction mechanism), resulting in a more balanced set of solutions depth-wise.

With this, we believe that our contribution is twofold. First, ideal mating preferences encoded as a second chromosome are not only sensible to the mutation method used (deeply affecting how preferences evolve), but they also seem to fail to reach a sustainable preference diversity on their own. Accordingly, this work gives us the notion that under these conditions the metaphor of mate choice is not self-sufficient, with its reliability being dependent on the set-up used. 
Second, and regardless of the first point, it still manages to serve its main purpose even when preference convergence occurs, driving the population through two different selective pressures and promoting more diverse solutions overall.

\section*{Declarations}
\begin{itemize}
\item \textbf{Funding} The first author is funded by national funds through the FCT - Foundation for Science and Technology, I.P., within the scope of the project CISUC - UI/BD/151046/2021 and by European Social Fund, through the Regional Operational Program Centro 2020. This work was partially supported by the Portuguese Recovery and Resilience Plan (PRR) through project C645008882-00000055, Center for Responsible AI, by the FCT, I.P./MCTES through national funds (PIDDAC), Project No. 7059 - Neuraspace - AI fights Space Debris, reference C644877546-00000020, and by the FCT - Foundation for Science and Technology, I.P./MCTES through national funds (PIDDAC), within the scope of CISUC R\&D Unit - UIDB/00326/2020 or project code UIDP/00326/2020.
\item \textbf{Conflict of interest} Not applicable.
\item \textbf{Ethics approval} Not applicable.
\item \textbf{Consent for publication} Not applicable.
\end{itemize}


\bibliography{sn-bibliography}

\end{document}